\pdfoutput=1

\documentclass[11pt]{article}

\usepackage[final]{acl}

\usepackage{times}
\usepackage{latexsym}

\usepackage[T1]{fontenc}

\usepackage[utf8]{inputenc}

\usepackage{microtype}

\usepackage{inconsolata}

\usepackage{graphicx}

\usepackage{xcolor}
\usepackage{tcolorbox}

\usepackage{lipsum}
\usepackage{geometry}

\usepackage{microtype}
\usepackage{url}
\usepackage{booktabs}

\usepackage{subfigure}
\usepackage{enumitem}
\newenvironment{itemize*}%
 {\leftmargini=20pt\begin{itemize}%
  \setlength{\itemsep}{3pt}%
  \setlength{\parskip}{0pt}%
  }%
 {\end{itemize}}
\newenvironment{enumerate*}%
 {\begin{enumerate}%
  \setlength{\itemsep}{0pt}%
  \setlength{\parskip}{0pt}}%
 {\end{enumerate}}

\usepackage{amsmath}
\usepackage{cleveref}
\usepackage{listings}
\usepackage{titlesec}
\usepackage{multirow}
\usepackage{makecell}

\usepackage{array}       
\usepackage{caption}     
\usepackage{amssymb}
\usepackage{pdfpages}

\definecolor{lightred}{RGB}{255,163,163}
\definecolor{deepred}{RGB}{146,0,0}
\definecolor{midnightgreen}{rgb}{0.0, 0.29, 0.33}
\definecolor{deepgreen}{HTML}{0aa344}
\definecolor{deeppurple}{HTML}{7030a0}
\definecolor{deepblue}{HTML}{171d91}
\definecolor{brown}{HTML}{843c0c}
\definecolor{shadered}{HTML}{ffe5e5}
\definecolor{shadegreen}{HTML}{e5f7ed}
\definecolor{teal}{HTML}{008080}
\definecolor{brown}{HTML}{8b4513}

\definecolor{skill_green}{HTML}{5c8d40}
\definecolor{skill_red}{HTML}{b02418}
\definecolor{skill_purple}{HTML}{6e276b}
\definecolor{skill_blue}{HTML}{4fadea}
\definecolor{skill_orange}{HTML}{da7842}

\newcommand{\deepred}{\textcolor{brown}}
\newcommand{\deepgreen}{\textcolor{teal}}

\usepackage{pifont}
\newcommand{\cmark}{\textcolor[rgb]{0.0, 0.6, 0.0}{\ding{51}}} 
\newcommand{\xmark}{\textcolor[rgb]{0.7, 0.0, 0.0}{\ding{55}}} 
\newcommand{\gmark}{\textcolor[rgb]{1,0.647,0}{\ding{51}}}

\NewDocumentCommand{\heng}
{ mO{} }{\textcolor{red}{\textsuperscript{\textit{Heng}}\textsf{\textbf{\small[#1]}}}}
\NewDocumentCommand{\cheng}
{ mO{} }{\textcolor{orange}{\textsuperscript{\textit{Cheng}}\textsf{\textbf{\small[#1]}}}}
\NewDocumentCommand{\xiusi}
{ mO{} }{\textcolor{purple}{\textsuperscript{\textit{Xiusi}}\textsf{\textbf{\small[#1]}}}}
\NewDocumentCommand{\hongru}
{ mO{} }{\textcolor{cyan}{\textsuperscript{\textit{Hongru}}\textsf{\textbf{\small[#1]}}}}
\NewDocumentCommand{\avi}
{ mO{} }{\textcolor{blue}{\textsuperscript{\textit{Avi}}\textsf{\textbf{\small[#1]}}}}
\NewDocumentCommand{\jiaxin}
{ mO{} }{\textcolor{pink}{\textsuperscript{\textit{Jiaxin}}\textsf{\textbf{\small[#1]}}}}
\NewDocumentCommand{\kmnote}
{ mO{} }{\textcolor{deepgreen}{\textsuperscript{\textit{Kathy}}\textsf{\textbf{\small[#1]}}}}

\usepackage[normalem]{ulem}        

\usepackage{algorithm}
\usepackage{algpseudocode}

\title{ModelingAgent: Bridging LLMs and Mathematical Modeling for Real-World Challenges}

\author{
Cheng Qian$^{1}$, Hongyi Du$^{1}$, Hongru Wang$^{1}$, Xiusi Chen$^{1}$, Yuji Zhang$^{1}$, Avirup Sil$^{2}$,\\
\textbf{Chengxiang Zhai$^{1}$, Kathleen McKeown$^{3}$, Heng Ji$^{1}$}\\
$^{1}$University of Illinois Urbana-Champaign, $^{2}$IBM Research AI, $^{3}$Columbia University\\
\texttt{\{chengq9, xiusic, hengji\}@illinois.edu}\\
}

\begin{document}
\maketitle
\begin{abstract}
Recent progress in large language models (LLMs) has enabled substantial advances in solving mathematical problems. However, existing benchmarks often fail to reflect the complexity of real-world problems, which demand open-ended, interdisciplinary reasoning and integration of computational tools. To address this gap, we introduce \textbf{ModelingBench}, a novel benchmark featuring real-world-inspired, open-ended problems from math modeling competitions across diverse domains, ranging from urban traffic optimization to ecosystem resource planning. These tasks require translating natural language into formal mathematical formulations, applying appropriate tools, and producing structured, defensible reports. ModelingBench also supports multiple valid solutions, capturing the ambiguity and creativity of practical modeling.
We also present \textbf{ModelingAgent}, a multi-agent framework that coordinates tool use, supports structured workflows, and enables iterative self-refinement to generate well-grounded, creative solutions. To evaluate outputs, we further propose \textbf{ModelingJudge}, an expert-in-the-loop system leveraging LLMs as domain-specialized judges assessing solutions from multiple expert perspectives. Empirical results show that ModelingAgent substantially outperforms strong baselines and often produces solutions indistinguishable from those of human experts. Together, our work provides a comprehensive framework for evaluating and advancing real-world problem-solving in open-ended, interdisciplinary modeling challenges.
All the codes are publicly released to facilitate future research~\footnote{\url{https://github.com/qiancheng0/ModelingAgent}}.
\end{abstract}

\section{Introduction}




\begin{figure*}
    \centering
    \includegraphics[width=1.0\linewidth]{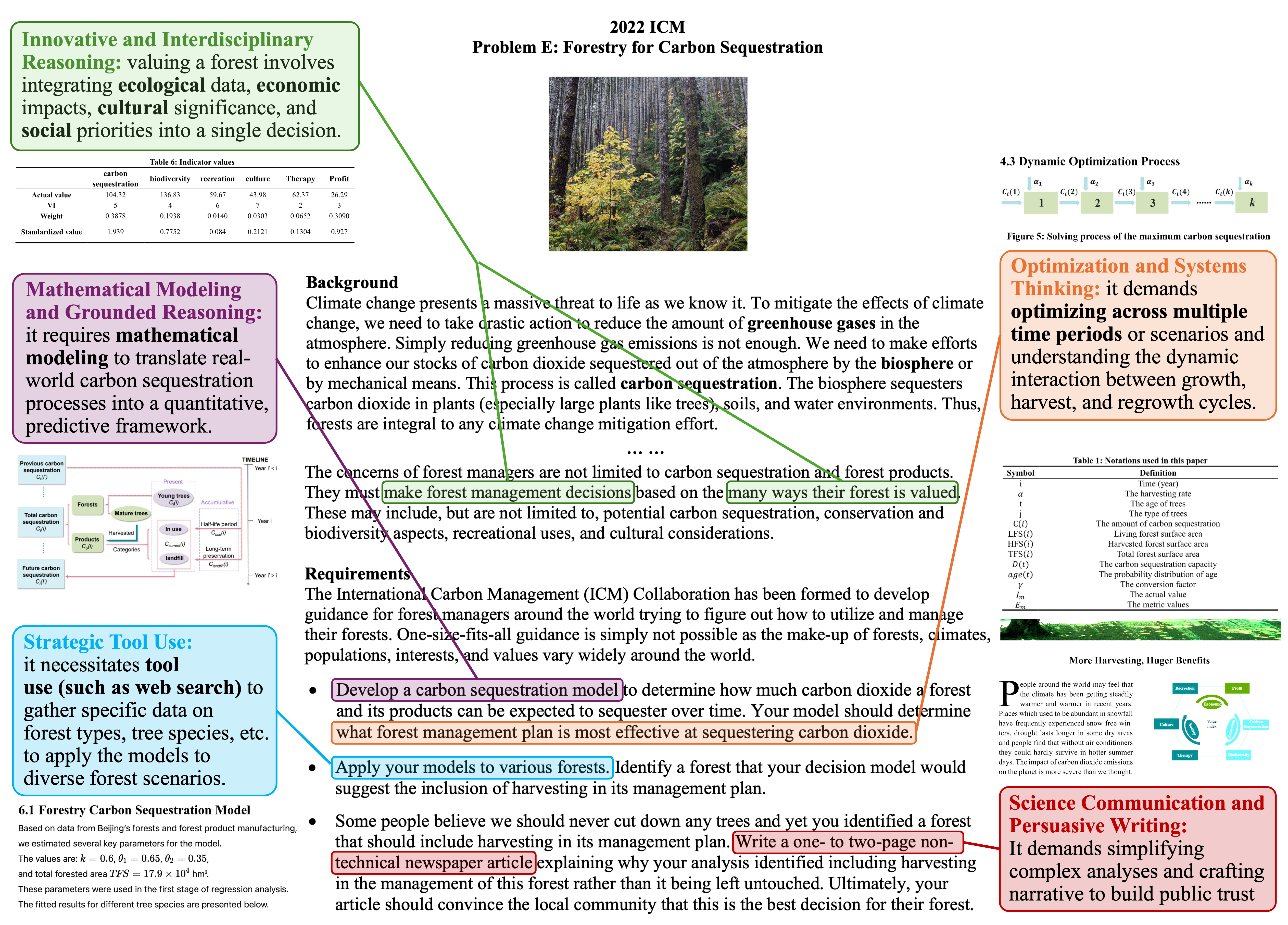}
    \caption{An example math modeling problem and the five core corresponding skills required.}
    \label{fig:skill}
    \vspace{-5pt}
\end{figure*}



Understanding and navigating the real world is a hallmark of human intelligence \citep{bassett2011human}. At its core, intelligence is not merely about retrieving facts or manipulating symbols, but about perceiving complex, often ambiguous situations and making sound, goal-directed decisions. One of the most powerful tools humans have developed for this purpose is mathematics: not just for abstract puzzles, but to structure messy, dynamic scenarios into analyzable forms \citep{giordano2013first}. This process—\textbf{Mathematical Modeling}—is fundamental to human reasoning in science, economics, and policy-making \citep{craddock2025strategic}. It involves translating real-world situations into formal mathematical representations, analyzing them, and interpreting the results to inform decisions \citep{guhhc2024modeling}.

{\small
\begin{tcolorbox}[colback=gray!5!white, colframe=black!75!black,
title=Definition of Mathematical Modeling, boxrule=0.3mm, width=0.48\textwidth, arc=1.8mm, auto outer arc=true]
Mathematical Modeling is the process of formulating an abstract model in terms of mathematical language to describe the complex behavior of a real system.
\end{tcolorbox}
}


In this sense, mathematical modeling is not just a technical skill but a testbed for real-world problem-solving intelligence.
Recent advances in LLMs have shown impressive performance on abstract mathematical problems such as symbolic algebra, theorem proving and puzzle solving, but often fall short on tasks grounded in the real world \citep{satpute2024llms}. Existing benchmarks typically emphasize well-posed, decontextualized problems, with little attention to contextual understanding or domain-specific considerations. For instance, a model may compute an integral or prove a lemma, but fail to model disease spread under resource constraints~\citep{shah2024infectious, sujau2025disease} or design a cost-effective transport network \citep{jonnala2024using, long2025survey}. These are the kinds of problems where modeling is central. Solving them requires more than computation: it demands rigorous mathematical formulation to ground reasoning in reality, strategic use of tools for active data acquisition, and an interdisciplinary perspective to foster innovative solutions. Developing benchmarks and models that engage with such tasks is essential to push LLMs toward more grounded and practically useful intelligence.



To bridge this gap, we introduce \textbf{ModelingBench}, a novel benchmark designed to evaluate LLMs on real-world modeling problems. Unlike existing evaluations, ModelingBench presents open-ended tasks that require holistic problem understanding, flexible tool use, data-driven analysis, and creative modeling strategies. As illustrated in \Cref{fig:skill}, these skills are essential for solving complex modeling problems\footnote{Find the full problem here: \url{https://www.mathmodels.org/Problems/2022/ICM-E/2022_ICM_Problem_E.pdf}}. The benchmark spans diverse domains, including sports analytics, financial modeling, biological systems, and operations management, thus encouraging interdisciplinary reasoning and innovation. An illustration of problem categories is further provided in \Cref{tab:examples}. While all tasks center on mathematical modeling, their real-world grounding makes ModelingBench a rigorous test of LLMs’ practical intelligence. In addition, inspired by the unrestricted tool access available to human participants, ModelingBench also offers a rich set of tools (detailed in \Cref{tab:tools}) including file operations, web access, and code execution, creating a sandbox environment for free exploration and end-to-end modeling.

\begin{table*}[!ht]
\centering
\resizebox{\linewidth}{!}{
\begin{tabular}{clcccccccc}
\toprule
\textbf{Dimension} & \textbf{Benchmark} & \textbf{\makecell{Tool\\Use}} & \textbf{\makecell{Environment\\Interaction}} & \textbf{\makecell{Math\\Reasoning}} & \textbf{\makecell{Math\\Modeling}} & \textbf{\makecell{Domain\\Expertise}} & \textbf{\makecell{Interdisci-\\plinarity}} & \textbf{Creativity} &\textbf{\makecell{Real-\\Worldness}} \\
\midrule

\multirow{4}{*}{\textbf{\makecell{Math \& \\Logic}}}
& \textit{OlymMATH\citep{sun2025challenging}}      
& \xmark & \xmark & \cmark & \xmark & \cmark & \xmark & \xmark & \xmark \\
& \textit{AIME\citep{AIME2025}}
& \xmark & \xmark & \cmark & \xmark & \cmark & \xmark & \xmark & \xmark \\
& \textit{AMC\citep{AMC2025}}      
& \xmark & \xmark & \cmark & \xmark & \cmark & \xmark & \xmark & \xmark \\
& \textit{GSM8K\citep{cobbe2021training}}      
& \xmark & \xmark & \cmark & \gmark & \xmark & \xmark & \xmark & \xmark \\

\midrule

\multirow{4}{*}{\textbf{\makecell{Expert\\Knowledge}}}
& \textit{PHYBench\citep{qiu2025phybench}}      
& \xmark & \xmark & \gmark & \gmark & \cmark & \xmark & \xmark & \xmark \\
& \textit{HLE\citep{phan2025humanity}}      
& \xmark & \xmark & \gmark & \xmark & \cmark & \xmark & \xmark & \xmark \\
& \textit{OlympiadBench\citep{he2024olympiadbench}}      
& \xmark & \xmark & \gmark & \xmark & \cmark & \xmark & \xmark & \xmark \\
& \textit{GPQA\citep{rein2024gpqa}}      
& \xmark & \xmark & \gmark & \xmark & \cmark & \xmark & \xmark & \xmark \\

\midrule

\multirow{4}{*}{\textbf{\makecell{Tool Use}}} 
& \textit{AppBench\citep{wang2024appbench}}      
& \cmark & \cmark & \xmark & \xmark & \xmark & \xmark & \xmark & \cmark \\
& \textit{TauBench\citep{yao2024tau}}      
& \cmark & \cmark & \xmark & \xmark & \xmark & \xmark & \xmark & \gmark \\
& \textit{BFCL\citep{patil2024gorilla}}      
& \cmark & \xmark & \xmark & \xmark & \xmark & \xmark & \xmark & \gmark \\
& \textit{ToolBench\citep{qin2023toolllm}}      
& \cmark & \xmark & \xmark & \xmark & \xmark & \xmark & \xmark & \cmark \\

\midrule

\multirow{4}{*}{\textbf{\makecell{Action\\Decision}}} 
& \textit{EmbodiedBench\citep{yang2025embodiedbench}}      
& \xmark & \cmark & \xmark & \xmark & \xmark & \xmark & \xmark & \xmark \\
& \textit{EscapeBench\citep{qian2024escapebench}}      
& \cmark & \cmark & \xmark & \xmark & \xmark & \xmark & \cmark & \xmark \\
& \textit{EA Interface\citep{li2024embodied}}      
& \xmark & \cmark & \xmark & \xmark & \xmark & \xmark & \xmark & \xmark \\
& \textit{Med Triage\citep{hu2024language}}      
& \xmark & \xmark & \xmark & \gmark & \cmark & \xmark & \xmark & \cmark \\

\midrule

\multirow{1}{*}{\textbf{\makecell{Ours}}} 
& \textit{ModelingBench}      
& \cmark & \cmark & \cmark & \cmark & \cmark & \cmark & \cmark & \cmark \\

\bottomrule
\end{tabular}
}
\caption{For each existing benchmark, the table indicates whether the corresponding ability is fully addressed (\cmark), partially addressed (\gmark), or not addressed (\xmark).}
\label{tab:comparison}
\end{table*}

\begin{table*}[!ht]
\centering
\scriptsize
\tabcolsep=0.005\linewidth
\begin{tabular}{llll}
\toprule
\textbf{Category} & \textbf{Year \& Contest} & \textbf{Title} & \textbf{Description} \\
\midrule
\textbf{Public Health} & 2019 MCM & The Opioid Crisis & \makecell[l]{Analyze multi-state drug report data to model opioid usage, focusing on \\narcotic analgesics and heroin through statistical and geographical modeling.} \\
\midrule
\textbf{Emergency Services} & 2013 HiMCM & Emergency Medical Response & \makecell[l]{Optimize ambulance placement across six zones to maximize 8-minute \\response coverage, incorporating scenarios including disasters.} \\
\midrule
\textbf{Smart Technology} & 2018 HiMCM & Cozy Smart House & \makecell[l]{Design a smart climate control system that adapts to irregular schedules \\and environmental factors using AI/ML and energy optimization.} \\
\midrule
\textbf{Environmental Engineering} & 2017 MCM & Managing The Zambezi River & \makecell[l]{Evaluate management strategies for the Kariba Dam, considering \\safety, environmental impact, and water flow across the Zambezi River.} \\
\bottomrule
\end{tabular}
\caption{Example problems in the ModelingBench of different categories.}
\label{tab:examples}
\end{table*}

\begin{figure*}
    \centering
    \includegraphics[width=1.0\linewidth]{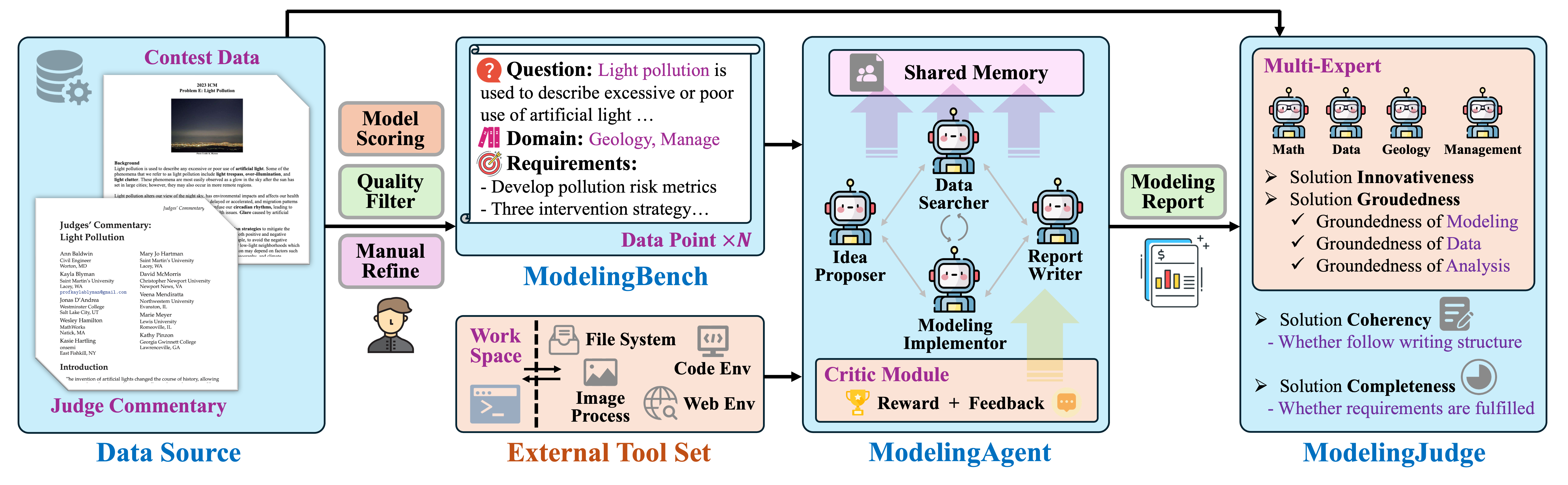}
    \caption{The automated system for solving and evaluating modeling problems.}
    \label{fig:pipeline}
\end{figure*}

To further support complex and grounded problem solving, we introduce \textbf{ModelingAgent}, a multi-agent framework composed of four key roles: an \textit{Idea Proposer}, a \textit{Data Searcher}, a \textit{Modeling Implementor}, and a \textit{Report Writer}. These agents share a common memory space and collaborate to automatically build rigorous mathematical models. The framework also incorporates a \textit{Critic Module} that evaluates and refines the modeling workflow from multiple perspectives, enabling self-improvement and optimization in an agentic, iterative fashion.


Given the open-ended, goal-oriented nature of modeling problems, we design our evaluation to mirror the assessment criteria used in real-world MCM competitions based on official judges' commentary. Specifically, we emphasize the completeness, structural coherence, and quality of final modeling reports, with a particular focus on solution groundedness and innovativeness. To address the subjectivity inherent in evaluating open-ended modeling tasks, we also introduce \textbf{ModelingJudge}, a multi-role LLM-based framework where models assume the roles of math experts, data experts, and problem-specific evaluators. This setup simulates real-world expert-in-the-loop grading practices while automating the evaluation of complex, open-ended problems.

Our experimental results show that ModelingAgent significantly outperforms strong baselines equipped with planning, reasoning, and free-form tool use, achieving up to a 20\% absolute improvement. However, a performance gap of around 10\% remains compared to award-winning human solutions, indicating room for further improvement in structural coherence, solution completeness, and analytical depth. We also analyze the critic module’s scoring behavior, which shows a clear upward trend over time, validating the effectiveness and transparency of ModelingAgent as a self-improving framework. Additionally, human evaluations reveal that the model’s implementations successfully fooled human judges over 50\% of the time, further demonstrating its ability to produce convincing, human-like solutions. In summary, our contributions are threefold:
\begin{itemize}[topsep=2pt, partopsep=-5pt, leftmargin=8pt, itemsep=-4pt]
\item We propose \textit{ModelingBench}, the first benchmark presenting mathematical modeling as a test of LLMs' real-world intelligence through open-ended real-world challenges.
\item We introduce \textit{ModelingAgent}, a multi-agent system inspired by real-world human collaboration, featuring a generalizable self-evolution algorithm that enables iterative improvement.
\item We develop \textit{ModelingJudge}, a competition-aligned, LLM-based evaluation framework that enables expert-in-the-loop automatic judging for open-ended modeling tasks.
\end{itemize}
As LLMs approach saturation on standard math benchmarks, we offer this work as a foundation for more practical, grounded, and interpretable evaluations of LLM's real-world intelligence.

\section{Related Work}

\paragraph{Evaluation on LLM Agent Intelligence.}
Sternberg’s Triarchic Theory of Intelligence~\cite{sternberg1997triarchic} divides intelligence into three components: analytical, practical, and creative. When adapted to the context of LLM agents, analytical intelligence corresponds to an agent’s foundational reasoning abilities, which can be assessed through reliable reasoning skills~\cite{wang2024rethinking, putta2024agent, zhang2025law} and effective tool use~\cite{wu2023autogen, liu2023agentbench, liu2024llava} across domains such as mathematics~\citep{cobbe2021training, AIME2025}, question answering~\citep{yang2018hotpotqa}, and planning~\citep{xie2024travelplanner}. Practical intelligence emphasizes the agent’s adaptability in real-world scenarios, demonstrated through proactive interactions with various environments~\citep{lu2024proactive}, including tool environments~\citep{li2023api}, web environments~\citep{yao2022webshop, zhou2023webarena}, embodied environments~\citep{li2024embodied, yang2025embodiedbench}, and game environments~\citep{costarelli2024gamebench}. Finally, creative intelligence, the least explored dimension, challenges agents to develop novel solutions~\citep{qian2023creator, cai2023large} and engage in creative tool use~\citep{qian2024escapebench} for efficient problem-solving. Building on this framework, our ModelingBench provides a comprehensive testbed for evaluating all three types of intelligence, with a focus on math reasoning, grounded real-world interactions, and innovative modeling strategies.

\paragraph{Collaborative Agents for Real-World Problem Solving}
Our work focuses on the practical application of multi-agent systems to real-world problems, a domain where LLM agents often face significant challenges~\citep{huang2025llms}. Prior studies have explored grounded multi-agent applications in various fields, including legal contract review~\citep{li2024legalagentbench}, academic writing~\citep{gao2025reviewagents}, code generation~\citep{zhu2025multiagentbench}, and scientific experiment automation~\citep{ghafarollahi2024sciagents}. While sharing the emphasis on collaboration, our approach specifically targets complex mathematical modeling tasks. Furthermore, our framework incorporates continuous self-improvement mechanisms inspired by agent reflection and self-evolution strategies~\citep{de2023emergent, qian2024investigate}. Previous self-improvement methods have primarily focused on memory enhancement~\citep{guo2023empowering, hatalis2023memory, zhang2024survey} and continual learning~\citep{majumder2023clin, dai2025multi} to support long-term context retention and incremental knowledge updates. Building on these, we introduce a critic module that enables self-feedback and solution scoring, substantially improving the real-world groundedness of agent-generated solutions.

\section{ModelingBench}

\begin{table*}[!ht]
\centering
\scriptsize
\tabcolsep=0.01\linewidth
\resizebox{\linewidth}{!}{
\begin{tabular}{@{\hspace{10pt}} ll @{\hspace{30pt}}}
\toprule
\multicolumn{2}{c}{\textbf{File Operations}} \\
\midrule
File Reader & Reads and processes various file formats (json, csv, txt), returning structured text with 50,000 character limit \\
\midrule
File Writer & Writes or appends content to files in write or append mode \\
\midrule
File Lister & Lists all files in a given directory recursively \\
\midrule
File Extractor & Extracts various compressed file formats (zip, tar, tar.gz, tar.bz2, gz, bz2, rar, 7z) \\
\addlinespace[1.5pt]
\toprule
\multicolumn{2}{c}{\textbf{Web Operations}} \\
\midrule
Web Search & Performs web searches using an API and returns structured search results \\
\midrule
Web Download & Downloads files from URLs and saves them to specified locations \\
\midrule
URL Extractor & Extracts all text content from a given webpage URL \\
\addlinespace[1.5pt]
\toprule
\multicolumn{2}{c}{\textbf{Image and Document Operations}} \\
\midrule
Image Captioner & Generates detailed captions for images using OpenAI's multimodal model \\
\midrule
Text Detector & Detects and extracts text from images using EasyOCR with multi-language support \\
\midrule
PDF Parser & Extracts and processes text from PDF documents with page selection options \\
\addlinespace[1.5pt]
\toprule
\multicolumn{2}{c}{\textbf{Other Tools}} \\
\midrule
Python Execution & Executes Python code from files or provided content with error handling \\
\midrule
Solution Generator & General-purpose tool that generates responses to queries, including image analysis \\
\bottomrule
\end{tabular}
}
\caption{Tool categories and functions in the sandbox environment for the modeling task.}
\label{tab:tools}
\end{table*}

\paragraph{Motivation.}
Mathematical modeling stands at the intersection of theory and real-world application, transforming abstract mathematical principles into tools for understanding, simulating, and solving complex problems. Historically rooted in practical necessity, mathematics has long served as a lens through which we interpret and navigate the world \citep{giordano2013first}. Modeling continues this by requiring not only computational competence but also creativity, domain knowledge, and strategic thinking. It offers a compelling framework for evaluating intelligence, as it engages core cognitive skills such as problem abstraction, data interpretation, and adaptive reasoning.

\paragraph{Data Source.}
To construct a benchmark that authentically tests these grounded and multifaceted abilities, we draw inspiration from the extensive suite of international modeling contests organized by COMAP\footnote{\url{https://www.comap.com/contests}}. These contests are widely recognized for promoting problem-solving and modeling excellence at various educational levels, including:
\begin{itemize}[topsep=2pt, partopsep=-5pt, leftmargin=8pt]
    \item \textbf{MCM/ICM}: The Mathematical and Interdisciplinary Contests in Modeling for undergraduate students, emphasizing continuous, discrete, and interdisciplinary problem domains.
    \item \textbf{HiMCM/MidMCM}: High school and middle school level contests focused on accessible yet realistic modeling tasks.
    \item \textbf{IM$^2$C}: The International Mathematical Modeling Challenge, fostering global engagement in real-world modeling.
\end{itemize}
All problems from these contests originate from real-world or policy-motivated challenges, validated by interdisciplinary panels of educators and experts. We collect all problem settings from the publicly available online database\footnote{\url{https://www.mathmodels.org}}, covering contest years from 2000 to 2025. These problems span a wide range of domains, including environmental science, public policy, epidemiology, and social systems, offering a rich and diverse foundation for constructing a comprehensive and interdisciplinary modeling benchmark.

\begin{table}[!t]
\centering
\label{tab:modelingbench-stats}
\resizebox{1.0\linewidth}{!}{
\begin{tabular}{l l l}
\toprule
\textbf{Total Problems} & \textbf{Avg. Subtasks} & \textbf{Domains Covered} \\
68 & 7.31 & 70+ \\
\midrule
\textbf{Domain Examples} & \multicolumn{2}{l}{\makecell{Epidemiology, Environmental Science, \\ Sports, Emergency Management, etc.}} \\
\midrule
\textbf{Year Span} & \multicolumn{2}{l}{2000 -- 2025} \\
\midrule
\textbf{Source} & \multicolumn{1}{l|}{\textbf{Count}} & \textbf{Difficulty} \\
MidMCM & \multicolumn{1}{l|}{3} & Easy: 6 \\
HiMCM & \multicolumn{1}{l|}{19} & Medium: 38 \\
MCM & \multicolumn{1}{l|}{24} & Hard: 24 \\
ICM & \multicolumn{1}{l|}{20} & -- \\
IM$^2$C & \multicolumn{1}{l|}{2} & -- \\ 
\bottomrule
\end{tabular}
}
\caption{Statistics of the \textit{ModelingBench} problems.}
\label{tab:statistics}
\end{table}

\paragraph{Data Filtering.}
As illustrated on the left of \Cref{fig:pipeline}, to ensure the quality and feasibility of our benchmark, we begin by using the GPT-4o model to heuristically rate over all the problems across three key dimensions: \textit{data accessibility}, \textit{modeling difficulty}, and \textit{image clarity} (See \Cref{apdx:data_curation} for details). Guided by these scores, we then conduct a manual filtering process to retain only high-quality problems that meet the following criteria: (1) the required data is either readily available online, making it accessible to LLMs equipped with web search tools, or is directly provided within the problem description; (2) the modeling task is feasible for LLMs to perform, without requiring physical actions such as measuring distances on a map or interacting physically with the real world; and (3) if images are included as essential information sources, they are clear enough to be accurately converted into textual descriptions, ensuring compatibility with text-only models equipped with multi-modal tools.

After this rigorous filtering and refinement, we finalize a curated subset of 68 high-quality problems of three difficulty levels for inclusion in ModelingBench from the original pool of 100+ problems. Please refer to \Cref{tab:tools} for detailed statistics.

\paragraph{Tool Augmentation.}
In real-world competitions, participants are free to use computers for tasks such as information searching and coding, as they are essential for building well-grounded models and conducting thorough analyses. To ensure that LLMs can similarly tackle problems effectively in our benchmark, we provide an augmented \textit{sandbox environment} equipped with a comprehensive suite of tools. This includes a core workspace where the model can perform arbitrary operations, a file management system for reading and writing files, web search capabilities for retrieving real-world information, and image processing tools for extracting information from multi-modal inputs. Additionally, we offer a collection of commonly used functionalities, such as code execution and PDF parsing, wrapped as callable tools to support complex reasoning and analysis. Please refer to \Cref{tab:tools} for a detailed list of available tools and explanations.


This sandbox environment constitutes the operational setting for our agent design. Within this space, models are free to interact with tools, interpret the problem, and autonomously simulate a real-world modeling workflow from scratch.


Overall, through this construction, ModelingBench ensures authenticity, diversity, and challenge, making it a powerful benchmark for assessing LLMs' real-world problem-solving abilities in math modeling contexts.


\section{ModelingAgent}


\begin{figure*}
    \centering
    \includegraphics[width=1.0\linewidth]{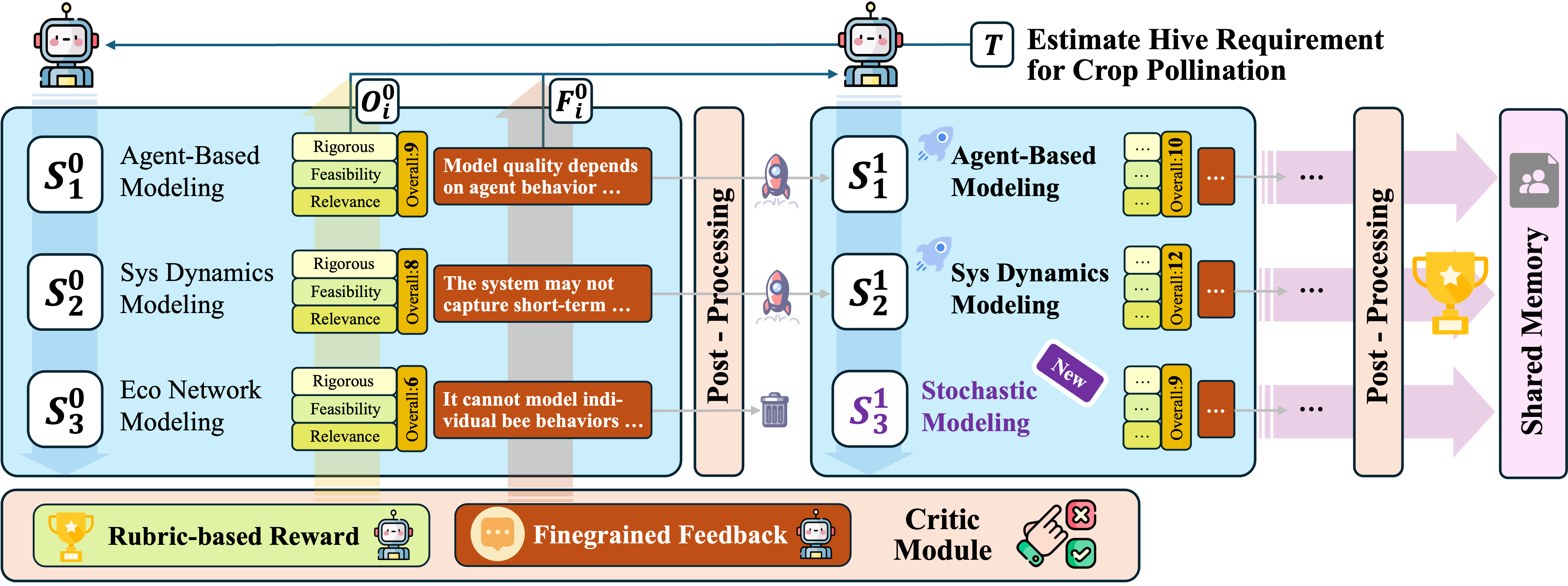}
    \caption{An illustration of iterative refinement performed by the critic module in ModelingAgent.}
    \label{fig:critic}
\end{figure*}


\paragraph{Motivation.}
To more effectively tackle mathematical modeling problems, we draw inspiration from the human team dynamics demonstrated in real-world competitions. In typical math modeling contests, teams are composed of three to four individuals with diverse areas of expertise, guided by a mentor who provides strategic advice and feedback. This collaborative structure naturally motivates a multi-agent framework designed to emulate such a problem-solving environment using LLMs.

\subsection{Multi-Agent Framework}
\label{sec:multi-agent_framework}
Based on the five core skills identified in \Cref{fig:skill}, we introduce a multi-agent framework composed of four specialized agents: Idea Proposer \(A_{IP}\), Data Searcher \(A_{DS}\), Model Implementor \(A_{MI}\), and Report Writer\(A_{RW}\). Each agent is designed to tackle complex mathematical modeling tasks through iterative interactions, coordinated and refined by a central critic module \(C\). All agents communicate via a shared memory, enabling seamless information exchange and collective problem-solving.

\paragraph{Idea Proposer \(A_{IP}\) (\textit{w.r.t.} \textcolor{skill_green}{Innovative and Interdisciplinary Reasoning}).} 
The Idea Proposer is responsible for generating suitable modeling approaches tailored to the given problem (\(T_{A_{IP}}\)). To accomplish this goal, \(A_{IP}\) is instructed to: (1) decompose the problem into clear, manageable subtasks; and (2) abstract and simplify these subtasks while providing explicit justifications. For each subtask, it proposes initial modeling ideas and iteratively refines them through interaction with the critic module \(C\). While \(A_{IP}\) may refer to a predefined list of common modeling techniques detailed in \Cref{apdx:reference_model}, it is encouraged to creatively adapt methods to the context of the problem.



\paragraph{Data Searcher \(A_{DS}\) (\textit{w.r.t.} \textcolor{skill_blue}{Strategic Tool Use}).} 
The Data Searcher is responsible for locating real-world datasets to support the implementation of the proposed model (\(T_{A_{DS}}\)). To accomplish this goal, \(A_{DS}\) is instructed to: (1) identify key variables required by the modeling approach, and (2) actively leverage available tools within the sandbox environment. During the data search process, it continuously interacts with the sandbox and refines its trajectory through feedback from the critic module \(C\). Importantly, \(A_{DS}\) closely engages external web resources to ensure that the resulting models are grounded in authentic and reliable data.



\paragraph{Model Implementor \(A_{MI}\) (\textit{w.r.t.} \textcolor{skill_purple}{Mathematical Modeling and Grounded Reasoning}).} 
The Model Implementor is responsible for transforming abstract modeling ideas into precise, executable mathematical formulations (\(T_{A_{MI}}^1\)), and implementing these models in code to generate results and conduct analysis (\(T_{A_{MI}}^2\)). To fulfill these objectives, \(A_{MI}\) is instructed to: (1) translate conceptual proposals into rigorous mathematical expressions, and (2) utilize the provided tools to instantiate the models empirically and analyze the outcomes. Throughout the process, \(A_{MI}\) interacts with the critic module \(C\) to iteratively refine both the mathematical formulations and their computational implementations.



\paragraph{Report Writer \(A_{RW}\) (\textit{w.r.t.} \textcolor{skill_red}{Science Communication and Persuasive Writing}).} 
The Report Writer is responsible for synthesizing the activities of all other agents into a coherent and comprehensive final report. Specifically, \(A_{RW}\) is instructed to: (1) dynamically identify and retrieve relevant information from the shared memory, and (2) organize the overall modeling workflow into a well-structured narrative. This agent continuously interacts with all modules and the shared memory, integrating their outputs into a polished report that serves as the final deliverable for evaluation.



\paragraph{Critic Module \(C\) (\textit{w.r.t.} \textcolor{skill_orange}{Optimization and Systems Thinking}).} 
The Critic Module plays a central role by interacting with all other agents. It is solely responsible for providing feedback and scoring each agent's behavior \(A\) based on its specific goal \(T_{A}\). Acting as a ``mentor,'' the critic guides the ``student'' agents through iterative feedback, helping them improve their performance and coordination. The detailed design and refinement algorithm of the critic module is presented in \Cref{sec:critic_design}.

\paragraph{Shared Memory.} The shared memory serves as a central hub for information exchange, playing a crucial role in coordinating interactions between agents. Conceptually, it can be viewed as an enhanced version of a scratch pad, offering more structured and organized information management. It is implemented as a dictionary, where each key encodes both the source of the information (i.e., which agent provided it) and the nature of the content. Agents store information by generating unique keys and retrieve it using the corresponding identifiers. This design not only enables flexible and efficient information access but also facilitates collaboration among agents, ultimately supporting the coherent assembly of the final report.



\subsection{Multi-Agent Orchestration}
As shown in the center of \Cref{fig:pipeline}, each agent can read from and write to this memory, enabling seamless coordination. All task trajectories (e.g., data search, model implementation) and interactions with the critic module \(C\) are automatically recorded in memory.

The modeling process begins with \(A_{IP}\), which receives the initial modeling task and proposes candidate modeling ideas, storing them in memory (an illustrative example is shown in \Cref{fig:case_study}). Based on these proposals, \(A_{DS}\) and \(A_{MI}\) collaborate closely to develop each subtask. Specifically, \(A_{MI}\) formalizes the modeling ideas into precise mathematical formulations, which guide \(A_{DS}\) in identifying the real-world data needed for modeling. Conversely, the data retrieved by \(A_{DS}\) may influence or constrain how \(A_{MI}\) implements the model and analyzes the results. Both agents iteratively refine their outputs through interactions with \(C\), with all updates recorded in the shared memory.

Simultaneously, \(A_{RW}\) monitors the memory and composes the final modeling report, synthesizing outputs from all agents into a coherent and comprehensive solution. In our experiments, all four agents and the critic module are powered by the same underlying language model to ensure fair and consistent evaluation. Full prompting strategies and configurations are detailed in \Cref{apdx:modeling_agent}.


\subsection{Critic Module Design}
\label{sec:critic_design}


{\small
\begin{algorithm}[htbp]
\small
\caption{{Iterative Solution Refinement}}
\label{algo:critic_refinement}
\begin{algorithmic}[1]
\Require Agent $A$, Critic $C$, Target $T_A$, Rubrics $\mathcal{R}_T$
\Require solution count $n$, discard count $k$, max iter $M$
\State Initialize solutions: $\mathcal{S}^{0}\sim A(\cdot|T_A)$
\For{$m=0,\dots,M-1$}
    \For{$S_i^{m}\in \mathcal{S}^{m}$}
        \State $O_i^{m}\gets\sum_{j} C(S_i^{m},R_T^{j})$
        \State $F_i^{m}\gets C(S_i^{m},\mathcal{R}_T)$
    \EndFor
    \State Sort: $\mathcal{S}^{m}\gets\{S_1^{m},\dots,S_{n-k}^{m}\}, O_1^{m}\geq\dots\geq O_n^{m}$
    \State Refine: $S_i^{m+1}\sim A(\cdot|T,S_i^{m},F_i^{m})$, $i\in[1,n-k]$
    \State Explore: $S_j^{m+1}\sim A(\cdot|T)$, $j\in[n-k+1,n]$
    \State Update: $\mathcal{S}^{m+1}\gets\{S_1^{m+1},\dots,S_n^{m+1}\}$
\EndFor
\State Final: $S_{\text{final}}\gets\arg\max_{S_i^{M}\in\mathcal{S}^{M}} O_i^{M}$
\State\Return $S_{\text{final}}$
\label{algo:critic}
\end{algorithmic}
\end{algorithm}
}

The critic module is integrated within multiple agent workflows to enhance specific attributes of the target \(T\), which vary according to each agent's distinct goal. In this section, we present the algorithm for designing and implementing the critic module.

Our critic procedure is outlined in \Cref{algo:critic}. Suppose we have an agent \( A \) aiming to accomplish a specific target \( T_{A} \), and let the critic module be denoted by \( C \). Initially, the agent is instructed to generate a set of \( n \) candidate solutions:
\begin{equation*}
    \mathcal{S}^{0} = \{ S_{1}^{0}, S_{2}^{0}, \dots, S_{n}^{0} \} \sim A(\cdot \mid T_{A}),
\end{equation*}
where the superscript \( 0 \) indicates these solutions belong to the initial generation. 



Next, the critic module \( C \) evaluates each candidate solution according to a set of \( m \) rubrics that specifically pertain to the target \( T \). These rubrics are represented as: \( \mathcal{R}_T = \{ R_{T}^{1}, R_{T}^{2}, \dots, R_{T}^{m} \} \).
For each rubric \( R_{T}^{j} \), the critic assigns a subscore along with targeted feedback to help improve the evaluated solution. The overall evaluation score \( O_{i}^{0} \) is calculated as the sum of these subscores, and the combined feedback is represented as \( F_{i}^{0} \) for each solution \( S_{i}^{0} \):
\begin{equation*}
    O_{i}^{0} \sim \sum_{j=1}^{m} C(\cdot \mid S_{i}^{0}, R_{T}^{j}), \quad F_{i}^{0} \sim C(\cdot \mid S_{i}^{0}, \mathcal{R}_T)
\end{equation*}
Following the evaluation, solutions proceed to a post-processing phase. To ensure efficient allocation of computational resources and maintain solution quality, the critic discards the bottom \( k \) solutions based on their scores. Consequently, the top \( n - k \) solutions with the highest scores are retained.





To maintain a consistent pool of candidate solutions, the agent generates \( k \) new solutions. This step explicitly encourages innovative exploration by instructing the agent to generate novel solutions inspired by the critic's feedback. Hence, the solution set for the next iteration is composed of:
\begin{itemize}[topsep=2pt, partopsep=-5pt, leftmargin=8pt]
    \item Refined top solutions: \( S_{1}^{1}, S_{2}^{1}, \dots, S_{n - k}^{1} \), obtained by refining solutions based on feedback.
    \item Explored new solutions: \( S_{n - k + 1}^{1}, \dots, S_{n}^{1} \), replacing the previously discarded solutions.
\end{itemize}


This evaluation and refinement process iterates up to a predetermined maximum iteration count \( M \). After the final iteration, the set of solutions \( \mathcal{S}^{M} = \{ S_{1}^{M}, S_{2}^{M}, \dots, S_{n}^{M} \} \) undergoes a final selection phase. The highest-scoring solution is selected as the final candidate \( S_{\text{final}} \) for subsequent stages of the modeling pipeline:
\begin{equation*}
    S_{\text{final}} = \underset{S_{i}^{M} \in \mathcal{S}^{M}}{\arg\max}\; O_{i}^{M}.
\end{equation*}

An illustrative example of this critic process is shown in \Cref{fig:critic}, considering a scenario with \( n = 3 \) and \( k = 1 \). Initially, the agent proposes three distinct modeling approaches. Each solution receives feedback and scores from the critic module. The lowest-scoring solution (the Eco Network Model) is discarded and replaced by a newly introduced solution (the Stochastic Model) in the subsequent iteration, while the remaining solutions undergo further refinement based on the critic's feedback. After the final iteration, the whole trajectory and the final selected solution will be put into the shared memory.

It is important to note that the provided algorithm describes a general-purpose critic mechanism applicable across various agent-task scenarios. In our multi-agent system, this modular critic design is utilized repeatedly for diverse goals, for which we discussed in \Cref{sec:multi-agent_framework}.

\section{ModelingJudge}
\label{sec:modeling_judge}

\paragraph{Motivation.}
Given the open-ended nature of tasks in ModelingBench, we further introduce \textbf{ModelingJudge}—a multi-expert-in-the-loop evaluation framework designed to simulate real-world modeling competition settings. In contests such as MCM/ICM, team rankings are based solely on the quality of submitted reports. ModelingJudge mirrors this setup by evaluating LLM performance exclusively through the final modeling report, which must comprehensively reflect the modeling process and address all task requirements. Drawing inspiration from MCM's multi-judge review structure, our framework incorporates diverse expert perspectives to enable a more nuanced and robust evaluation.

\paragraph{Expert Role Incorporation.}
ModelingJudge uses LLMs to simulate multiple expert roles. Each evaluation includes a mathematical modeling expert and a data analysis expert, as these competencies are universally required across modeling problems. In addition, two domain-specific experts are selected based on the problem context provided in ModelingBench. Each expert is instructed to assess the report from their disciplinary perspective. For example, in the problem ``Forestry for Carbon Sequestration'' presented in \Cref{fig:skill}, an environmental scientist will focus on evaluating the model's treatment of ecological factors, such as biodiversity, soil health, and ecosystem services—drawing on principles from ecology, conservation biology, and environmental science.

\paragraph{Evaluation Metrics.}
We assess each report along three core dimensions, adapted from COMAP’s official judging commentary:
\begin{enumerate}[topsep=0pt, partopsep=-5pt, leftmargin=12pt, itemsep=-3pt]
    \item \textbf{Structural Coherency}: Clarity and organization of the report, including the presence of key sections such as assumptions, model formulation, solution process, and analysis.
    \item \textbf{Solution Completeness}: Whether the report addresses all sub-questions and task requirements defined in the ModelingBench problem.
    \item \textbf{Solution Quality}, which further includes:
    \begin{itemize}[topsep=2pt, partopsep=-5pt, leftmargin=8pt, itemsep=0pt]
        \item \textbf{Groundedness of Modeling}: Rigor, relevance, and appropriateness of the modeling techniques adapted to customized scenarios.
        \item \textbf{Groundedness of Data}: Authenticity, adequacy, and contextual relevance of the data applied for the modeling process.
        \item \textbf{Groundedness of Analysis}: Depth of analysis, correctness of mathematical reasoning, and interpretative insight.
        \item \textbf{Innovativeness}: Originality and potential real-world utility of the modeling approach.
    \end{itemize}
\end{enumerate}

Given the inherent subjectivity of solution quality, we leverage the diverse perspectives of multiple expert roles to ensure balanced and fair evaluation. In contrast, structural coherency and solution completeness are assessed using a single LLM-as-judge configuration, as these dimensions are relatively more objective and consistent across tasks. Details on evaluation prompts and configurations can be found in \Cref{apdx:modeling_judge}.



\section{Experiments}






In this section, we present benchmarking results on ModelingBench and evaluate the effectiveness of the ModelingAgent framework in addressing complex modeling problems.

\subsection{Experiment Setup}

\paragraph{Baselines.}
We compare \textbf{ModelingAgent} against two primary baselines:
(1) \textit{Vanilla Generation}, where the model is instructed to directly generate a mathematical modeling report from the prompt without access to any tools;
(2) \textit{Tool Agent}, where the model is granted full access to the sandbox environment and equipped with a planner to autonomously decide how to use tools to solve the problem, serving as a strong agent-style baseline.
These two baselines also serve as ablation variants: the first tests performance without tool access, and the second without structured, role-based guidance. All instruction details are in \Cref{apdx:prompt_details}.

\paragraph{Models.}
We evaluate a range of open-source and closed-source LLMs across multiple families, including GPT-4o\citep{hurst2024gpt}, Deepseek-Chat~\citep{liu2024deepseekV3}, Gemini-2.0-Flash, Gemini-2.0-Thinking~\citep{team2023gemini}, Llama3.1-72B-Instruct~\citep{dubey2024llama}, Qwen2.5-70B-Instruct~\citep{2024qwen2.5}, and QwQ-32B~\citep{qwen2024qwq}. Our evaluation includes not only LLMs but also Large Reasoning Models (LRMs) to comprehensively assess the effectiveness of our method. We exclude smaller-scale models (around 7B parameters), as they empirically struggle with following complex instructions and providing the final modeling report.

\paragraph{Evaluation Metrics.}
We use the ModelingJudge framework and the scoring criteria described in \Cref{sec:modeling_judge}. For solution quality, scores from different expert roles are averaged. All reported results are then averaged across the full set of problems in the ModelingBench. While our framework allows for weighted combinations of metrics, the final score reported (in the last column) is currently computed as a simple average, with equal weight assigned to each evaluation metric.

\begin{table*}[t]
    \centering
    \setlength\tabcolsep{2pt}
    \setlength\extrarowheight{2pt}
    
    \resizebox{\linewidth}{!}{
    \begin{tabular}{l >{\hspace{7pt}}c<{\hspace{10pt}} >{\hspace{10pt}}c<{\hspace{10pt}} >{\hspace{10pt}}c<{\hspace{7pt}} >{\hspace{7pt}}c<{\hspace{7pt}} >{\hspace{7pt}}c<{\hspace{7pt}} >{\hspace{7pt}}c<{\hspace{10pt}} c}

        \toprule

        \multirow{2}{*}{\textbf{Model}} & \multirow{2}{*}{\textbf{\makecell{Structural\\Coherence}}} & \multirow{2}{*}{\textbf{\makecell{Solution\\Completeness}}} & \multicolumn{4}{c}{\textbf{\makecell{Solution Quality}}} & \multirow{2}{*}{\textbf{\makecell{Average}}} \\
        
        \cmidrule(lr){4-7}
        
        ~ & ~ & ~ & \textbf{\makecell{Modeling Groundedness}} & \textbf{\makecell{Data Groundedness}} & \textbf{\makecell{Analysis Groundedness}} & \textbf{\makecell{Innovativeness}} & ~ \\
        
        \addlinespace[2pt]
        \midrule
        \addlinespace[2pt]
        \multicolumn{8}{c}{\textit{Vanilla Model Generation}} \\ 
        \midrule
        
        \textit{GPT-4o} & 75.00 & 55.80 & 53.33 & 41.71 & 49.10 & 29.63 & 50.76 \\
        \textit{Deepseek-Chat} & 79.78 & 64.52 & 59.01 & 44.82 & 52.81 & 35.94 & 56.14 \\
        \textit{Gemini-2.0-Flash} & 78.68 & 62.12 & 54.37 & 38.64 & 50.35 & 33.99 & 53.03 \\
        \textit{Gemini-2.0-Think} & 83.46 & 61.34 & 52.79 & 31.34 & 51.31 & 38.20 & 53.07 \\
        \textit{Llama3.1-70B-Instruct} & 63.24 & 42.12 & 44.89 & 24.65 & 42.10 & 23.62 & 40.10 \\
        \textit{Qwen2.5-72B-Instruct} & 61.62 & 44.42 & 44.01 & 29.93 & 40.74 & 25.07 & 40.96 \\
        \textit{QwQ-32B} & 69.26 & 55.67 & 49.43 & 31.80 & 47.43 & 41.18 & 49.13 \\
        
        \addlinespace[2pt]
        \midrule
        \addlinespace[2pt]
        \multicolumn{8}{c}{\textit{Base Tool Agent}} \\ 
        \midrule
        
        \textit{GPT-4o} & 73.24\deepred{$_{\downarrow 1.76}$} & 64.69\deepgreen{$_{\uparrow 8.89}$} & 55.57\deepgreen{$_{\uparrow 2.24}$} & 49.41\deepgreen{$_{\uparrow 7.70}$} & 51.07\deepgreen{$_{\uparrow 1.97}$} & 38.11\deepgreen{$_{\uparrow 8.48}$} & 55.35\deepgreen{$_{\uparrow 4.59}$} \\
        \textit{Deepseek-Chat} & 76.99\deepred{$_{\downarrow 2.79}$} & 67.50\deepgreen{$_{\uparrow 2.98}$} & 67.43\deepgreen{$_{\uparrow 8.42}$} & 60.55\deepgreen{$_{\uparrow 15.73}$} & 63.46\deepgreen{$_{\uparrow 10.65}$} & 45.48\deepgreen{$_{\uparrow 9.54}$} & \uline{63.57}\deepgreen{$_{\uparrow 7.43}$} \\
        \textit{Gemini-2.0-Flash} & 72.50\deepred{$_{\downarrow 6.18}$} & 65.73\deepgreen{$_{\uparrow 3.61}$} & 55.83\deepgreen{$_{\uparrow 1.46}$} & 47.00\deepgreen{$_{\uparrow 8.36}$} & 56.36\deepgreen{$_{\uparrow 6.01}$} & 41.34\deepgreen{$_{\uparrow 7.35}$} & 56.46\deepgreen{$_{\uparrow 3.43}$} \\
        \textit{Gemini-2.0-Think} & 75.81\deepred{$_{\downarrow 7.65}$} & 59.29\deepred{$_{\downarrow 2.05}$} & 55.48\deepgreen{$_{\uparrow 2.69}$} & 38.27\deepgreen{$_{\uparrow 6.93}$} & 60.15\deepgreen{$_{\uparrow 8.84}$} & 43.40\deepgreen{$_{\uparrow 5.20}$} & 55.40\deepgreen{$_{\uparrow 2.33}$} \\
        \textit{Llama3.1-70B-Instruct} & 60.96\deepred{$_{\downarrow 2.28}$} & 41.70\deepred{$_{\downarrow 0.42}$} & 47.67\deepgreen{$_{\uparrow 2.78}$} & 35.50\deepgreen{$_{\uparrow 10.85}$} & 43.22\deepgreen{$_{\uparrow 1.12}$} & 28.27\deepgreen{$_{\uparrow 4.65}$} & 42.88\deepgreen{$_{\uparrow 2.78}$} \\
        \textit{Qwen2.5-72B-Instruct} & 73.82\deepgreen{$_{\uparrow 12.20}$} & 70.24\deepgreen{$_{\uparrow 25.82}$} & 54.87\deepgreen{$_{\uparrow 10.86}$} & 46.67\deepgreen{$_{\uparrow 16.74}$} & 55.44\deepgreen{$_{\uparrow 14.70}$} & 40.06\deepgreen{$_{\uparrow 15.00}$} & 56.85\deepgreen{$_{\uparrow 15.89}$} \\
        \textit{QwQ-32B} & 93.75\deepgreen{$_{\uparrow 24.49}$} & 70.60\deepgreen{$_{\uparrow 14.93}$} & 58.51\deepgreen{$_{\uparrow 9.08}$} & 55.37\deepgreen{$_{\uparrow 23.57}$} & 62.04\deepgreen{$_{\uparrow 14.61}$} & 49.56\deepgreen{$_{\uparrow 8.38}$} & \textbf{64.97}\deepgreen{$_{\uparrow 15.84}$} \\

        
        \addlinespace[2pt]
        \midrule
        \addlinespace[2pt]
        \multicolumn{8}{c}{\textit{ModelingAgent (Our Method)}} \\ 
        \midrule

        \textit{GPT-4o} & 81.84\deepgreen{$_{\uparrow 6.84}$} & 81.68\deepgreen{$_{\uparrow 16.99}$} & 69.52\deepgreen{$_{\uparrow 13.95}$} & 63.57\deepgreen{$_{\uparrow 14.16}$} & 70.97\deepgreen{$_{\uparrow 19.90}$} & 70.90\deepgreen{$_{\uparrow 32.79}$} & 73.08\deepgreen{$_{\uparrow 17.73}$} \\
        \textit{Deepseek-Chat} & 88.75\deepgreen{$_{\uparrow 8.97}$} & 87.95\deepgreen{$_{\uparrow 20.45}$} & 92.11\deepgreen{$_{\uparrow 24.68}$} & 76.18\deepgreen{$_{\uparrow 15.63}$} & 74.86\deepgreen{$_{\uparrow 11.40}$} & 74.65\deepgreen{$_{\uparrow 29.17}$} & \textbf{82.42\deepgreen{$_{\uparrow 18.85}$}} \\
        \textit{Gemini-2.0-Flash} & 78.46\deepred{$_{\downarrow 0.22}$} & 74.52\deepgreen{$_{\uparrow 8.79}$} & 72.39\deepgreen{$_{\uparrow 16.56}$} & 63.38\deepgreen{$_{\uparrow 16.38}$} & 72.79\deepgreen{$_{\uparrow 16.43}$} & 73.14\deepgreen{$_{\uparrow 31.80}$} & 72.45\deepgreen{$_{\uparrow 15.99}$} \\
        \textit{Gemini-2.0-Think} & 65.88\deepred{$_{\downarrow 17.58}$} & 63.19\deepgreen{$_{\uparrow 1.85}$} & 56.20\deepgreen{$_{\uparrow 0.72}$} & 37.13\deepred{$_{\downarrow 1.14}$} & 70.13\deepgreen{$_{\uparrow 9.98}$} & 64.16\deepgreen{$_{\uparrow 20.76}$} & 59.45\deepgreen{$_{\uparrow 4.05}$} \\
        \textit{Llama3.1-70B-Instruct} & 74.34\deepgreen{$_{\uparrow 11.10}$} & 72.49\deepgreen{$_{\uparrow 30.37}$} & 67.50\deepgreen{$_{\uparrow 19.83}$} & 69.98\deepgreen{$_{\uparrow 34.48}$} & 69.21\deepgreen{$_{\uparrow 25.99}$} & 67.81\deepgreen{$_{\uparrow 39.54}$} & 70.22\deepgreen{$_{\uparrow 27.34}$} \\
        \textit{Qwen2.5-72B-Instruct} & 83.25\deepgreen{$_{\uparrow 9.43}$} & 90.45\deepgreen{$_{\uparrow 20.21}$} & 76.44\deepgreen{$_{\uparrow 21.57}$} & 78.31\deepgreen{$_{\uparrow 31.64}$} & 70.88\deepgreen{$_{\uparrow 15.44}$} & 69.00\deepgreen{$_{\uparrow 28.94}$} & 78.05\deepgreen{$_{\uparrow 21.20}$} \\
        \textit{QwQ-32B} & 83.73\deepred{$_{\downarrow 10.02}$} & 87.47\deepgreen{$_{\uparrow 16.87}$} & 83.15\deepgreen{$_{\uparrow 24.64}$} & 77.13\deepgreen{$_{\uparrow 21.76}$} & 70.65\deepgreen{$_{\uparrow 8.61}$} & 71.39\deepgreen{$_{\uparrow 21.83}$} & \uline{78.92}\deepgreen{$_{\uparrow 13.95}$} \\


        \addlinespace[2pt]
        \midrule
        \addlinespace[2pt]
        \multicolumn{8}{c}{\textit{Top Human Report (Award Winning)}} \\ 
        \midrule
        
        \textit{Human Expert} & 96.30\deepgreen{$_{\uparrow 2.55}$} & 98.83\deepgreen{$_{\uparrow 10.88}$} & 80.32\deepred{$_{\downarrow 11.79}$} & 76.36\deepred{$_{\downarrow 0.77}$} & 85.31\deepgreen{$_{\uparrow 10.45}$} & 70.65\deepred{$_{\downarrow 4.00}$} & \textbf{84.63}\deepgreen{$_{\uparrow 2.21}$} \\
        
        \bottomrule
    \end{tabular}
    }
    \caption{Main results comparing the performance of different models across three settings: Vanilla Model Generation, Base Tool Agent, and ModelingAgent. The arrows indicate the increase or decrease relative to the \textit{highest} value in the corresponding position from the previous results.}
    \label{tab:result_main}
    \vspace{-8pt}
\end{table*}


\subsection{Results}
We present the main results in \Cref{tab:result_main}, highlighting the following key findings:

\paragraph{ModelingAgent effectively addresses modeling challenges.}  
Under the ModelingJudge framework, ModelingAgent outperforms both Vanilla Generation and Tool Agent by up to 20\%, demonstrating its superior capability in tackling mathematical modeling challenges. Notably, the solution quality score (including both groundedness and innovativeness) shows the largest improvement, with certain increases reaching up to 30\%. This improvement likely stems from the inclusion of an idea proposer, which enhances the diversity and creativity of high-level solution concepts. Additionally, the structured coordination in ModelingAgent promotes better-grounded analysis and reports, contributing to higher final scores.

\paragraph{Top human reports still outperform.}
Since our benchmark is based on real competitions, we compare ModelingAgent with award-winning human reports and find it still lags behind. This indicates room for further improvement. Humans achieve notably higher scores in both solution completeness and structural coherency, suggesting that LLMs still struggle to fully address the wide range of subtasks and complex requirements inherent in modeling tasks. Moreover, the flexibility and effectiveness of tool use for data collection and analysis remain inferior to human capabilities. We further analyze the differences between human and LLM performance through evaluations conducted by human judges in \Cref{sec:exp_analysis}. Interestingly, we also find that LRM’s performance does not significantly surpass that of LLMs, reinforcing that the challenges of modeling are universally difficult and that even models with enhanced reasoning still face similar limitations.

\paragraph{Innovativeness remains a challenge for LLMs.}  
From Vanilla Generation to Tool Agent, the average final score shows only marginal improvement; however, groundedness scores consistently increase by approximately 10\%, indicating that access to external tools enables LLMs to produce more data-driven and well-supported solutions. Upon upgrading to ModelingAgent, the innovativeness score improves substantially. Nevertheless, across all evaluation metrics, innovativeness consistently remains the lowest-scoring dimension, regardless of the method used. This highlights a fundamental challenge: LLMs still find it difficult to generate truly creative and human-level \textit{intelligent} solutions.


\begin{figure*}[!t]
    \centering
    \includegraphics[width=0.9\linewidth]{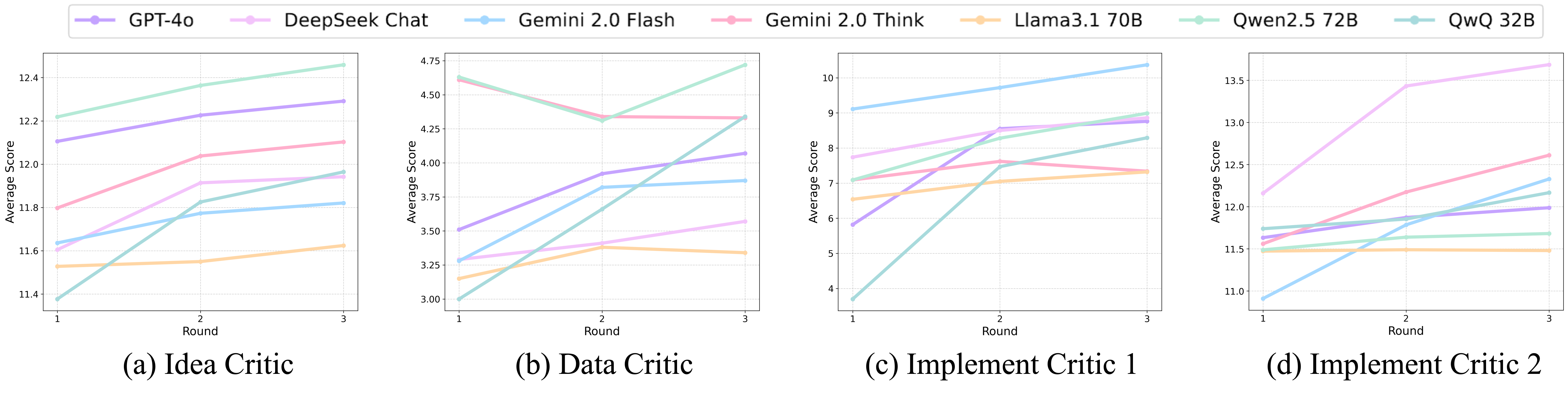}
    \caption{The critic’s scoring trend across rounds shows a clear upward trajectory, highlighting ModelingAgent’s consistent improvements and effective self-evolution in addressing modeling challenges.}
    \label{fig:combined_critic}
\end{figure*}


\subsection{Analysis}
\label{sec:exp_analysis}

\paragraph{Critic Trend Analysis.} \Cref{fig:combined_critic} shows the critic’s scoring trend over multiple agent-critic refinement rounds, covering idea proposal, data search, and model implementation. The consistent upward trend demonstrates both the effectiveness of ModelingAgent’s self-evolution in addressing modeling challenges and the critic’s adaptive evaluation based on agent performance. Note that absolute scores across models are not directly comparable due to potential scoring bias, but the continuous improvement within each model highlights the promise of multi-agent self-evolution inspired by human practices.

\begin{figure*}[t]
\centering
\small

\resizebox{1.0\textwidth}{!}{
\begin{tcolorbox}[colback=gray!5!white, colframe=blue!75!black, 
title=Case Study: Risk Assessment Models (From 2024 ICM - GPT-4o Solution), boxrule=0.3mm, width=\textwidth, arc=3mm, auto outer arc=true]

\textbf{Subtask:} Resource Requirements and Feasibility Analysis

\textbf{Objective:} Determine the resources required for the project and assess its feasibility.

\rule{\linewidth}{0.4pt}

\textbf{Round 1} \vspace{5pt}

\textbf{Approach:} \textit{Risk Assessment Models} \\
\textbf{Application:} Risk assessment models identify and evaluate potential risks associated with the project, such as financial, operational, or strategic risks. They help in developing risk mitigation strategies to enhance project feasibility.

\vspace{3pt} \textbf{Critic Feedback:}\\
$\bullet$ \textbf{Relevance:} Highly relevant for identifying risks and barriers; complements financial analysis. \textbf{(Score: 5)}\\
$\bullet$ \textbf{Rigor:} Uses probabilistic methods and scenario analysis; dependent on data quality. \textbf{(Score: 4)}\\
$\bullet$ \textbf{Practicality:} Feasible with access to risk data but resource-intensive. \textbf{(Score: 3)}\\
$\bullet$ \textbf{Overall Score: 12}

\rule{\linewidth}{0.4pt}

\textbf{Round 2} \vspace{5pt}

\textbf{Approach:} \textit{Risk Assessment Models with Monte Carlo Simulations} \\
\textbf{Enhanced Application:} Building upon the initial approach, the model now integrates Monte Carlo simulations to quantify the impact of various risk factors on project outcomes more accurately. This probabilistic approach enables decision-makers to visualize a range of possible outcomes under different risk scenarios, providing deeper insights into potential barriers and mitigation strategies.

\vspace{3pt} \textbf{Critic Feedback:} \\
$\bullet$ \textbf{Relevance:} Crucial for feasibility analysis; offers a structured approach to manage complex risks. \textbf{(Score: 5)}\\
$\bullet$ \textbf{Rigor:} Incorporates mathematically robust simulations and scenario analysis, enhancing model depth. \textbf{(Score: 5)}\\
$\bullet$ \textbf{Practicality:} Implementation is more complex, requiring specialized tools and expertise, but provides valuable insights, especially for high-stakes projects. \textbf{(Score: 4)}\\
$\bullet$ \textbf{Overall Score: 14}\\

\end{tcolorbox}
}

\caption{Case study of iterative refinement of modeling idea and corresponding critics.}
\label{fig:case_study}
\end{figure*}

\paragraph{Case Study.} As illustrated in \Cref{fig:case_study}, GPT-4o under our ModelingAgent framework initially applied standard risk assessment models to identify key risks. However, critics noted insufficient quantitative depth for informed decision-making. In direct response to this feedback, the model’s application was enhanced by integrating Monte Carlo simulations, enabling probabilistic analysis of risk impacts and offering more precise strategic insights. Recognizing this improvement, the critics revised their evaluation by increasing the overall score. This case highlights how feedback-driven model enhancement directly influenced critic's assessment, validating the iterative refinement.


\begin{figure*}[!t]
    \centering
    \includegraphics[width=0.95\linewidth]{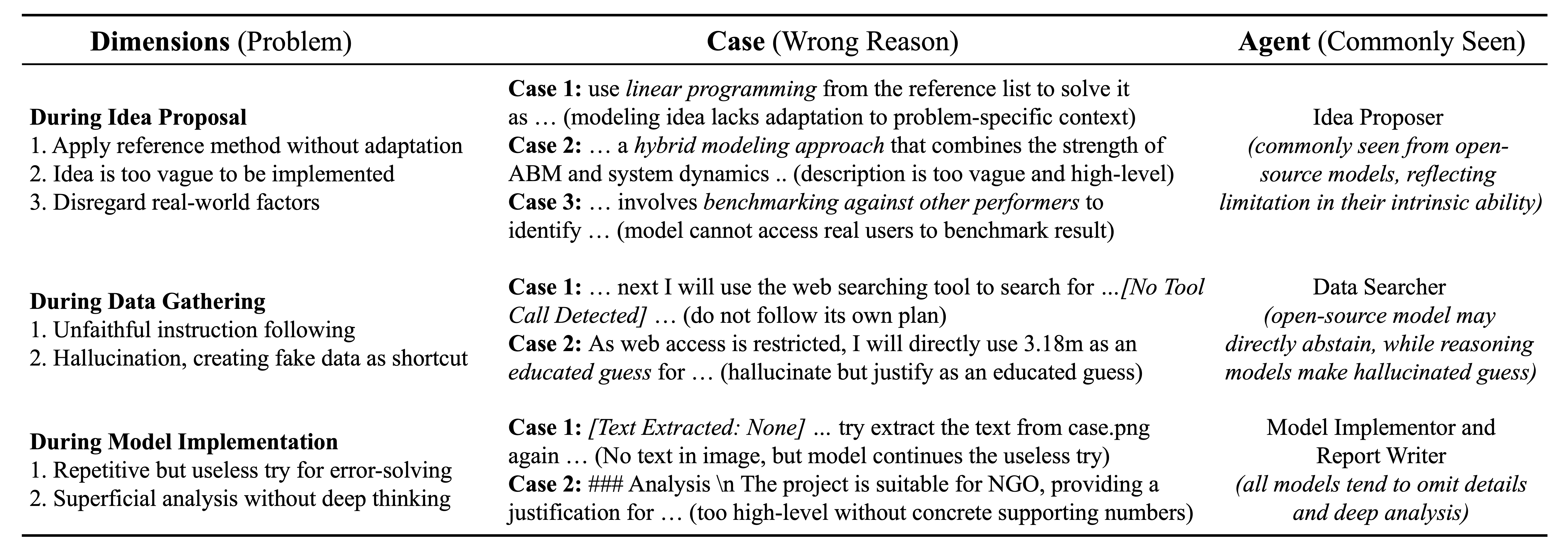}
    \caption{Summarization of common errors or imperfect cases presented throughout ModelingAgent framework.}
    \label{fig:error_analysis}
\end{figure*}

\paragraph{Error Analysis.} As shown in \Cref{fig:error_analysis}, the observed errors highlight that current models still often lack problem-specific reasoning, reliable data handling, and in-depth analysis capabilities. To further improve, models could be equipped with better uncertainty awareness to avoid hallucination, enhanced context handling and understanding mechanisms to propose actionable ideas and avoid lost in the middle, and stronger reasoning capability to support more rigorous and detailed analyses rather than relying on vague justifications.


\begin{figure}[!t]
    \centering
    \begin{minipage}{\linewidth}
        \centering
        \caption*{(a) Model Ranking}
        \resizebox{\linewidth}{!}{
            \begin{tabular}{lll}
            \toprule
            \textbf{Model Name}       & \textbf{Top Rank (\%)} & \textbf{Second Rank (\%)} \\
            \midrule
            GPT-4o                    & 8.33                  & 8.33                     \\
            Deepseek-Chat             & 4.17                  & 16.67                     \\
            Gemini-2.0-Flash          & 20.83                 & 8.33                     \\
            Gemini-2.0-Think          & 0.00                  & 12.50                     \\
            Llama3.1-70B-Instruct     & 0.00                  & 8.33                     \\
            Qwen2.5-72B-Instruct      & 16.67                 & \textbf{37.50}                     \\
            QwQ-32B                   & \textbf{50.00}        & 8.33                     \\
            \midrule
            \textbf{Total}            & 100.00                & 100.00                     \\
            \bottomrule
            \end{tabular}
        }
    \end{minipage}

    \vspace{15pt}
    
    \begin{minipage}{\linewidth}
        \centering
        \caption*{(b) Method Ranking}
        \resizebox{\linewidth}{!}{
            \begin{tabular}{lll}
            \toprule
            \textbf{Model Name}       & \textbf{Top Rank (\%)} & \textbf{Second Rank (\%)} \\
            \midrule
            Vanilla Model Generation  & 0.00              & 25.00                \\
            Base Tool Agent           & 12.50             & 8.33                 \\
            ModelingAgent             & \textbf{45.83}    & \textbf{41.67}       \\
            Human Expert Solution     & 41.67             & 25.00                \\
            \midrule
            \textbf{Total}            & 100.00            & 100.00               \\
            \bottomrule
            \end{tabular}
        }
    \end{minipage}

    \vspace{15pt}
    
    \begin{minipage}{0.85\linewidth}
        \centering
        \caption*{(c) Turing Test Result}
        \includegraphics[width=\linewidth]{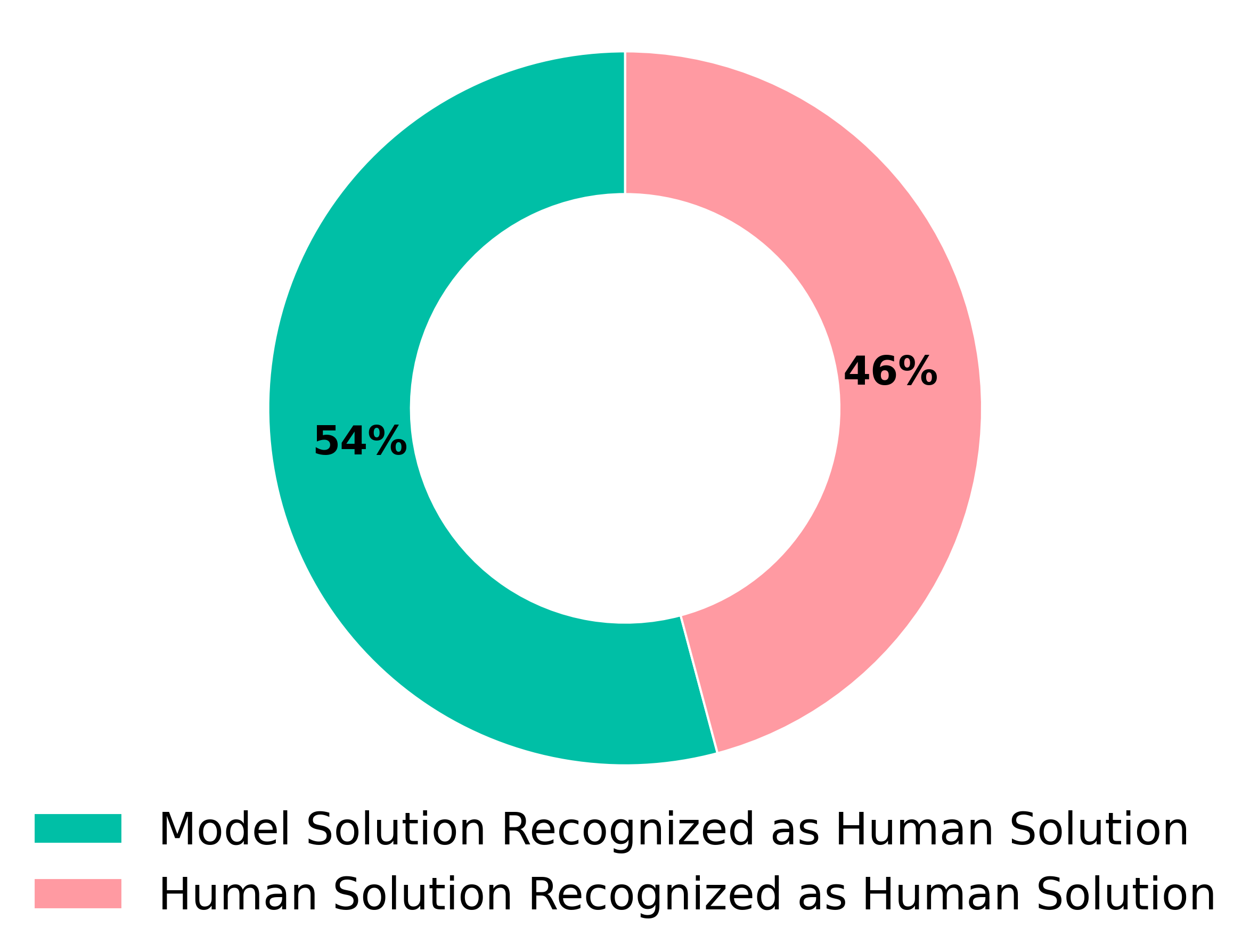}
    \end{minipage}
    \caption{Human evaluation results identifying the top-ranked and second-ranked modeling solutions based on (a) different models and (b) different methods. (c) The Turing test applied to model-generated solutions.}
    \label{fig:human_evaluation}
\end{figure}

\paragraph{Human Evaluation.} We conduct a human evaluation to assess the quality of the models’ math modeling reports. Recognizing the subjective nature of human judgments, we adopt an arena-style evaluation. Specifically, we randomly select modeling implementation writings for the same question and ask human evaluators to rank their quality. Detailed evaluation settings are provided in \Cref{apdx:human_eval}. This evaluation aims to answer three key questions: (1) Which model performs best? (2) Which method yields the best results? (3) Can the top human solution be distinguished from model-generated ones (i.e., a Turing Test)?

We present our human evaluation results in \Cref{fig:human_evaluation}. Under the ModelingAgent framework, we observe that solutions generated by QwQ-32B are significantly preferred by human evaluators. This aligns with the results in \Cref{tab:result_main}, where QwQ-32B also achieves the high average score. In contrast, the low-scoring Gemini-2.0-Think and Llama3.1-70B models are consistently less preferred and is never ranked as the top solution by participant, further validating the effectiveness of ModelingJudge and its alignment with human preferences.

Additionally, we discover the ModelingAgent’s solutions are significantly more favored over the other two baselines. Remarkably, even when compared to human expert modeling reports, ModelingAgent’s outputs are more frequently ranked as the top solution. This demonstrates the effectiveness of our approach. In Turing Test, over 50\% of the time, model-generated solutions were indistinguishable from top human solutions, highlighting that our method can produce reports comparable to human experts in both perceived quality and content.

\section{Conclusion and Future Work}
Our work introduces ModelingBench, a math modeling benchmark bridging abstract mathematical reasoning with real-world problem-solving, and ModelingAgent, a multi-agent LLM framework that supports complex, interdisciplinary modeling through structured collaboration, iterative refinement, and strategic tool use. Our ModelingJudge evaluation framework further enables real-world competition inspired, expert-aligned assessment. Together, these contributions demonstrate the potential of LLMs to address practical challenges in multiple domains. Despite clear improvements over baselines, limitations remain in creativity, data reliability, and domain adaptation, highlighting key areas for future research at the intersection of NLP, modeling, and real-world decision-making.

Future work could focus on expanding multi-modal reasoning capabilities, enabling models to seamlessly integrate visual, textual, and structured data essential for tackling complex, real-world problems in areas like climate resilience, healthcare, and economic policy. Additionally, advancing the agentic self-evolution framework with stronger causal reasoning and meaningful human-in-the-loop feedback can be critical to improving solution reliability and fostering deeper, more accountable decision-making processes. We envision this work as a foundation for rethinking how we evaluate LLM capabilities as their performance converges on standard benchmarks, and for inspiring new interdisciplinary methods that amplify their real-world impact at the intersection of NLP, mathematical modeling, and high-stakes decision-making.


\section*{Limitations}
This work primarily investigates the capabilities of LLMs and LRMs in addressing real-world challenges through the lens of mathematical modeling. However, it does not comprehensively evaluate Vision-Language Models (VLMs), which are increasingly critical for tasks requiring visual perception, such as interpreting maps, charts, and complex visual data. While we integrate a multi-modal understanding tool to mitigate this limitation, it serves only as a stopgap and does not represent true native visual reasoning. Extending our evaluation to include VLMs represents an important direction for future research. Notably, during data curation, we excluded many problems requiring physical simulation or complex visual understanding, which remain beyond the capabilities of current LLMs.

Additionally, our benchmark includes a limited set of problems, primarily due to two factors: (1) rigorous quality control processes resulted in the exclusion of many unsuitable data points, and (2) our problem set is constrained by the availability of modeling challenges from COMAP competitions. These factors make large-scale dataset expansion labor-intensive and challenging. Nevertheless, similar to established competitions like AIME and AMC, our benchmark is designed to be dynamic, with new modeling problems incorporated as they are released annually. This ensures that the benchmark remains relevant and reflective of evolving real-world challenges.

Finally, the open-ended nature of modeling tasks makes objective evaluation particularly challenging, especially in the absence of scalable human-in-the-loop assessments. While our proposed ModelingJudge framework simulates expert evaluations using LLMs and ensures consistent comparisons by employing GPT-4o across all experiments, potential biases and arbitrariness in automated judgment remain. To address this, we complement our evaluations with human user studies. We hope future work will build on this foundation to develop more robust, transparent, and unbiased automatic evaluation frameworks for open-ended, interdisciplinary problem-solving.

\section*{Ethical Statement}
We release ModelingBench and code for ModelingAgent solely for academic research purposes, with the aim of advancing the capabilities of LLMs in mathematical modeling and real-world decision-making. All problems included in our benchmark have undergone strict quality control and human supervision to ensure relevance, accuracy, and fairness. We strongly discourage any misuse or harmful application of this dataset.

While our benchmark addresses high-stakes societal challenges, we emphasize that model-generated solutions should be interpreted with caution and must undergo rigorous human oversight before being applied in real-world decision-making contexts. Additionally, although our ModelingAgent framework provides a powerful approach for solving complex modeling tasks, we caution against its use in competitive settings where the use of AI-generated content is restricted. Users must adhere to competition rules and ethical guidelines, avoiding any form of system misuse.

Ultimately, our goal is to promote transparency, fairness, and explainability in the modeling process, and to inspire the development of next-generation evaluation method that push LLMs toward more trustworthy and impactful real-world applications.

\bibliography{custom}

\begin{thebibliography}{59}
\providecommand{\natexlab}[1]{#1}

\bibitem[{Bassett and Gazzaniga(2011)}]{bassett2011human}
Danielle~S Bassett and Michael~S Gazzaniga. 2011.
\newblock Human intelligence and brain networks.
\newblock \emph{Dialogues in Clinical Neuroscience}, 13(4):395--401.

\bibitem[{Cai et~al.(2024)Cai, Wang, Ma, Chen, and Zhou}]{cai2023large}
Tianle Cai, Xuezhi Wang, Tengyu Ma, Xinyun Chen, and Denny Zhou. 2024.
\newblock \href {https://openreview.net/forum?id=qV83K9d5WB} {Large language models as tool makers}.
\newblock In \emph{The Twelfth International Conference on Learning Representations}.

\bibitem[{Cobbe et~al.(2021)Cobbe, Kosaraju, Bavarian, Chen, Jun, Kaiser, Plappert, Tworek, Hilton, Nakano et~al.}]{cobbe2021training}
Karl Cobbe, Vineet Kosaraju, Mohammad Bavarian, Mark Chen, Heewoo Jun, Lukasz Kaiser, Matthias Plappert, Jerry Tworek, Jacob Hilton, Reiichiro Nakano, et~al. 2021.
\newblock Training verifiers to solve math word problems.
\newblock \emph{arXiv preprint arXiv:2110.14168}.

\bibitem[{Costarelli et~al.(2024)Costarelli, Allen, Hauksson, Sodunke, Hariharan, Cheng, Li, Clymer, and Yadav}]{costarelli2024gamebench}
Anthony Costarelli, Mat Allen, Roman Hauksson, Grace Sodunke, Suhas Hariharan, Carlson Cheng, Wenjie Li, Joshua Clymer, and Arjun Yadav. 2024.
\newblock Gamebench: Evaluating strategic reasoning abilities of llm agents.
\newblock \emph{arXiv preprint arXiv:2406.06613}.

\bibitem[{Craddock(2025)}]{craddock2025strategic}
Michael Craddock. 2025.
\newblock \href {https://medium.com/@mcraddock/a-comprehensive-overview-of-mathematical-models-for-strategic-decision-making-f62bbbe7e4da} {A comprehensive overview of mathematical models for strategic decision-making}.

\bibitem[{Dai et~al.(2025)Dai, Xie, Liu, Wang, Li, Wang, and Lui}]{dai2025multi}
Xiangxiang Dai, Yuejin Xie, Maoli Liu, Xuchuang Wang, Zhuohua Li, Huanyu Wang, and John Lui. 2025.
\newblock Multi-agent conversational online learning for adaptive llm response identification.
\newblock \emph{arXiv preprint arXiv:2501.01849}.

\bibitem[{De~Zarz{\`a} et~al.(2023)De~Zarz{\`a}, De~Curt{\`o}, Roig, Manzoni, and Calafate}]{de2023emergent}
I~De~Zarz{\`a}, J~De~Curt{\`o}, Gemma Roig, Pietro Manzoni, and Carlos~T Calafate. 2023.
\newblock Emergent cooperation and strategy adaptation in multi-agent systems: An extended coevolutionary theory with llms.
\newblock \emph{Electronics}, 12(12):2722.

\bibitem[{Dubey et~al.(2024)Dubey, Jauhri, Pandey, Kadian, Al-Dahle, Letman, Mathur, Schelten, Yang, Fan et~al.}]{dubey2024llama}
Abhimanyu Dubey, Abhinav Jauhri, Abhinav Pandey, Abhishek Kadian, Ahmad Al-Dahle, Aiesha Letman, Akhil Mathur, Alan Schelten, Amy Yang, Angela Fan, et~al. 2024.
\newblock The llama 3 herd of models.
\newblock \emph{arXiv preprint arXiv:2407.21783}.

\bibitem[{Gao et~al.(2025)Gao, Ruan, Gao, Liu, and Fu}]{gao2025reviewagents}
Xian Gao, Jiacheng Ruan, Jingsheng Gao, Ting Liu, and Yuzhuo Fu. 2025.
\newblock Reviewagents: Bridging the gap between human and ai-generated paper reviews.
\newblock \emph{arXiv preprint arXiv:2503.08506}.

\bibitem[{Ghafarollahi and Buehler(2024)}]{ghafarollahi2024sciagents}
Alireza Ghafarollahi and Markus~J Buehler. 2024.
\newblock Sciagents: Automating scientific discovery through bioinspired multi-agent intelligent graph reasoning.
\newblock \emph{Advanced Materials}, page 2413523.

\bibitem[{Giordano et~al.(2013)Giordano, Fox, and Horton}]{giordano2013first}
F.R. Giordano, W.P. Fox, and S.B. Horton. 2013.
\newblock \href {https://books.google.com/books?id=PYUWAAAAQBAJ} {\emph{A First Course in Mathematical Modeling}}.
\newblock Cengage Learning.

\bibitem[{Guhhc(2024)}]{guhhc2024modeling}
Chinna~Raja Guhhc. 2024.
\newblock \href {https://www.linkedin.com/pulse/mathematical-modeling-real-world-problem-solving-chinna-raja-guhhc} {Mathematical modeling in real-world problem solving}.

\bibitem[{Guo et~al.(2023)Guo, Li, Qi, Yang, Li, Feng, Zhang, and Xu}]{guo2023empowering}
Jing Guo, Nan Li, Jianchuan Qi, Hang Yang, Ruiqiao Li, Yuzhen Feng, Si~Zhang, and Ming Xu. 2023.
\newblock Empowering working memory for large language model agents.
\newblock \emph{arXiv preprint arXiv:2312.17259}.

\bibitem[{Hatalis et~al.(2023)Hatalis, Christou, Myers, Jones, Lambert, Amos-Binks, Dannenhauer, and Dannenhauer}]{hatalis2023memory}
Kostas Hatalis, Despina Christou, Joshua Myers, Steven Jones, Keith Lambert, Adam Amos-Binks, Zohreh Dannenhauer, and Dustin Dannenhauer. 2023.
\newblock Memory matters: The need to improve long-term memory in llm-agents.
\newblock In \emph{Proceedings of the AAAI Symposium Series}, volume~2, pages 277--280.

\bibitem[{He et~al.(2024)He, Luo, Bai, Hu, Thai, Shen, Hu, Han, Huang, Zhang et~al.}]{he2024olympiadbench}
Chaoqun He, Renjie Luo, Yuzhuo Bai, Shengding Hu, Zhen~Leng Thai, Junhao Shen, Jinyi Hu, Xu~Han, Yujie Huang, Yuxiang Zhang, et~al. 2024.
\newblock Olympiadbench: A challenging benchmark for promoting agi with olympiad-level bilingual multimodal scientific problems.
\newblock \emph{arXiv preprint arXiv:2402.14008}.

\bibitem[{Hu et~al.(2024)Hu, Ray, Leung, Summerville, Joy, Funk, and Basharat}]{hu2024language}
Brian Hu, Bill Ray, Alice Leung, Amy Summerville, David Joy, Christopher Funk, and Arslan Basharat. 2024.
\newblock Language models are alignable decision-makers: Dataset and application to the medical triage domain.
\newblock \emph{arXiv preprint arXiv:2406.06435}.

\bibitem[{Huang et~al.(2025)Huang, Shen, Hu, Gao, and Wang}]{huang2025llms}
Xuhan Huang, Qingning Shen, Yan Hu, Anningzhe Gao, and Benyou Wang. 2025.
\newblock Llms for mathematical modeling: Towards bridging the gap between natural and mathematical languages.
\newblock In \emph{Findings of the Association for Computational Linguistics: NAACL 2025}, pages 2678--2710.

\bibitem[{Hurst et~al.(2024)Hurst, Lerer, Goucher, Perelman, Ramesh, Clark, Ostrow, Welihinda, Hayes, Radford et~al.}]{hurst2024gpt}
Aaron Hurst, Adam Lerer, Adam~P Goucher, Adam Perelman, Aditya Ramesh, Aidan Clark, AJ~Ostrow, Akila Welihinda, Alan Hayes, Alec Radford, et~al. 2024.
\newblock Gpt-4o system card.
\newblock \emph{arXiv preprint arXiv:2410.21276}.

\bibitem[{Jonnala et~al.(2024)Jonnala, Liang, Yang, and Alsmadi}]{jonnala2024using}
Ramya Jonnala, Gongbo Liang, Jeong Yang, and Izzat Alsmadi. 2024.
\newblock Using large language models in public transit systems, san antonio as a case study.
\newblock \emph{arXiv preprint arXiv:2407.11003}.

\bibitem[{Li et~al.(2024{\natexlab{a}})Li, Chen, Yang, Ai, Jia, Liu, Lin, Wu, Yuan, Hu et~al.}]{li2024legalagentbench}
Haitao Li, Junjie Chen, Jingli Yang, Qingyao Ai, Wei Jia, Youfeng Liu, Kai Lin, Yueyue Wu, Guozhi Yuan, Yiran Hu, et~al. 2024{\natexlab{a}}.
\newblock Legalagentbench: Evaluating llm agents in legal domain.
\newblock \emph{arXiv preprint arXiv:2412.17259}.

\bibitem[{Li et~al.(2024{\natexlab{b}})Li, Zhao, Wang, Wang, Zhou, Srivastava, Gokmen, Lee, Li, Zhang et~al.}]{li2024embodied}
Manling Li, Shiyu Zhao, Qineng Wang, Kangrui Wang, Yu~Zhou, Sanjana Srivastava, Cem Gokmen, Tony Lee, Erran~Li Li, Ruohan Zhang, et~al. 2024{\natexlab{b}}.
\newblock Embodied agent interface: Benchmarking llms for embodied decision making.
\newblock \emph{Advances in Neural Information Processing Systems}, 37:100428--100534.

\bibitem[{Li et~al.(2023)Li, Zhao, Yu, Song, Li, Yu, Li, Huang, and Li}]{li2023api}
Minghao Li, Yingxiu Zhao, Bowen Yu, Feifan Song, Hangyu Li, Haiyang Yu, Zhoujun Li, Fei Huang, and Yongbin Li. 2023.
\newblock Api-bank: A comprehensive benchmark for tool-augmented llms.
\newblock In \emph{Proceedings of the 2023 Conference on Empirical Methods in Natural Language Processing}, pages 3102--3116.

\bibitem[{Liu et~al.(2024{\natexlab{a}})Liu, Feng, Xue, Wang, Wu, Lu, Zhao, Deng, Zhang, Ruan et~al.}]{liu2024deepseekV3}
Aixin Liu, Bei Feng, Bing Xue, Bingxuan Wang, Bochao Wu, Chengda Lu, Chenggang Zhao, Chengqi Deng, Chenyu Zhang, Chong Ruan, et~al. 2024{\natexlab{a}}.
\newblock Deepseek-v3 technical report.
\newblock \emph{arXiv preprint arXiv:2412.19437}.

\bibitem[{Liu et~al.(2024{\natexlab{b}})Liu, Cheng, Liu, Zhang, Li, Ren, Zou, Yang, Su, Zhu et~al.}]{liu2024llava}
Shilong Liu, Hao Cheng, Haotian Liu, Hao Zhang, Feng Li, Tianhe Ren, Xueyan Zou, Jianwei Yang, Hang Su, Jun Zhu, et~al. 2024{\natexlab{b}}.
\newblock Llava-plus: Learning to use tools for creating multimodal agents.
\newblock In \emph{European Conference on Computer Vision}, pages 126--142. Springer.

\bibitem[{Liu et~al.(2023)Liu, Yu, Zhang, Xu, Lei, Lai, Gu, Ding, Men, Yang, Zhang, Deng, Zeng, Du, Zhang, Shen, Zhang, Su, Sun, Huang, Dong, and Tang}]{liu2023agentbench}
Xiao Liu, Hao Yu, Hanchen Zhang, Yifan Xu, Xuanyu Lei, Hanyu Lai, Yu~Gu, Hangliang Ding, Kaiwen Men, Kejuan Yang, Shudan Zhang, Xiang Deng, Aohan Zeng, Zhengxiao Du, Chenhui Zhang, Sheng Shen, Tianjun Zhang, Yu~Su, Huan Sun, Minlie Huang, Yuxiao Dong, and Jie Tang. 2023.
\newblock \href {https://arxiv.org/abs/2308.03688} {Agentbench: Evaluating llms as agents}.
\newblock \emph{Preprint}, arXiv:2308.03688.

\bibitem[{Long et~al.(2025)Long, Liu, Cao, Ren, Ju, Fang, Zhu, Zhu, and Zhou}]{long2025survey}
Qingqing Long, Shuai Liu, Ning Cao, Zhicheng Ren, Wei Ju, Chen Fang, Zhihong Zhu, Hengshu Zhu, and Yuanchun Zhou. 2025.
\newblock A survey of large language models for traffic forecasting: Methods and applications.
\newblock \emph{Authorea Preprints}.

\bibitem[{Lu et~al.(2024)Lu, Yang, Qian, Chen, Luo, Wu, Wang, Cong, Zhang, Lin et~al.}]{lu2024proactive}
Yaxi Lu, Shenzhi Yang, Cheng Qian, Guirong Chen, Qinyu Luo, Yesai Wu, Huadong Wang, Xin Cong, Zhong Zhang, Yankai Lin, et~al. 2024.
\newblock Proactive agent: Shifting llm agents from reactive responses to active assistance.
\newblock \emph{arXiv preprint arXiv:2410.12361}.

\bibitem[{Majumder et~al.(2023)Majumder, Mishra, Jansen, Tafjord, Tandon, Zhang, Callison-Burch, and Clark}]{majumder2023clin}
Bodhisattwa~Prasad Majumder, Bhavana~Dalvi Mishra, Peter Jansen, Oyvind Tafjord, Niket Tandon, Li~Zhang, Chris Callison-Burch, and Peter Clark. 2023.
\newblock Clin: A continually learning language agent for rapid task adaptation and generalization.
\newblock \emph{arXiv preprint arXiv:2310.10134}.

\bibitem[{of~America~(MAA)(2025{\natexlab{a}})}]{AIME2025}
Mathematical~Association of~America~(MAA). 2025{\natexlab{a}}.
\newblock American invitational mathematics examination.

\bibitem[{of~America~(MAA)(2025{\natexlab{b}})}]{AMC2025}
Mathematical~Association of~America~(MAA). 2025{\natexlab{b}}.
\newblock American mathematics competitions.

\bibitem[{Patil et~al.(2024)Patil, Zhang, Wang, and Gonzalez}]{patil2024gorilla}
Shishir~G Patil, Tianjun Zhang, Xin Wang, and Joseph~E Gonzalez. 2024.
\newblock Gorilla: Large language model connected with massive apis.
\newblock \emph{Advances in Neural Information Processing Systems}, 37:126544--126565.

\bibitem[{Phan et~al.(2025)Phan, Gatti, Han, Li, Hu, Zhang, Zhang, Shaaban, Ling, Shi et~al.}]{phan2025humanity}
Long Phan, Alice Gatti, Ziwen Han, Nathaniel Li, Josephina Hu, Hugh Zhang, Chen Bo~Calvin Zhang, Mohamed Shaaban, John Ling, Sean Shi, et~al. 2025.
\newblock Humanity's last exam.
\newblock \emph{arXiv preprint arXiv:2501.14249}.

\bibitem[{Putta et~al.(2024)Putta, Mills, Garg, Motwani, Finn, Garg, and Rafailov}]{putta2024agent}
Pranav Putta, Edmund Mills, Naman Garg, Sumeet Motwani, Chelsea Finn, Divyansh Garg, and Rafael Rafailov. 2024.
\newblock Agent q: Advanced reasoning and learning for autonomous ai agents.
\newblock \emph{arXiv preprint arXiv:2408.07199}.

\bibitem[{Qian et~al.(2023)Qian, Han, Fung, Qin, Liu, and Ji}]{qian2023creator}
Cheng Qian, Chi Han, Yi~Fung, Yujia Qin, Zhiyuan Liu, and Heng Ji. 2023.
\newblock Creator: Tool creation for disentangling abstract and concrete reasoning of large language models.
\newblock In \emph{Findings of the Association for Computational Linguistics: EMNLP 2023}, pages 6922--6939.

\bibitem[{Qian et~al.(2024{\natexlab{a}})Qian, Han, Luo, He, Chen, Zhang, Du, Yao, Yang, Zhang et~al.}]{qian2024escapebench}
Cheng Qian, Peixuan Han, Qinyu Luo, Bingxiang He, Xiusi Chen, Yuji Zhang, Hongyi Du, Jiarui Yao, Xiaocheng Yang, Denghui Zhang, et~al. 2024{\natexlab{a}}.
\newblock Escapebench: Pushing language models to think outside the box.
\newblock \emph{arXiv preprint arXiv:2412.13549}.

\bibitem[{Qian et~al.(2024{\natexlab{b}})Qian, Liang, Qin, Ye, Cong, Lin, Wu, Liu, and Sun}]{qian2024investigate}
Cheng Qian, Shihao Liang, Yujia Qin, Yining Ye, Xin Cong, Yankai Lin, Yesai Wu, Zhiyuan Liu, and Maosong Sun. 2024{\natexlab{b}}.
\newblock Investigate-consolidate-exploit: A general strategy for inter-task agent self-evolution.
\newblock \emph{arXiv preprint arXiv:2401.13996}.

\bibitem[{Qin et~al.(2024)Qin, Liang, Ye, Zhu, Yan, Lu, Lin, Cong, Tang, Qian, Zhao, Hong, Tian, Xie, Zhou, Gerstein, Li, Liu, and Sun}]{qin2023toolllm}
Yujia Qin, Shihao Liang, Yining Ye, Kunlun Zhu, Lan Yan, Yaxi Lu, Yankai Lin, Xin Cong, Xiangru Tang, Bill Qian, Sihan Zhao, Lauren Hong, Runchu Tian, Ruobing Xie, Jie Zhou, Mark Gerstein, Dahai Li, Zhiyuan Liu, and Maosong Sun. 2024.
\newblock Toolllm: Facilitating large language models to master 16000+ real-world apis.
\newblock In \emph{The Twelfth International Conference on Learning Representations}.

\bibitem[{Qiu et~al.(2025)Qiu, Guo, Song, Sun, Cai, Wei, Luo, Yin, Zhang, Hu et~al.}]{qiu2025phybench}
Shi Qiu, Shaoyang Guo, Zhuo-Yang Song, Yunbo Sun, Zeyu Cai, Jiashen Wei, Tianyu Luo, Yixuan Yin, Haoxu Zhang, Yi~Hu, et~al. 2025.
\newblock Phybench: Holistic evaluation of physical perception and reasoning in large language models.
\newblock \emph{arXiv preprint arXiv:2504.16074}.

\bibitem[{Rein et~al.(2024)Rein, Hou, Stickland, Petty, Pang, Dirani, Michael, and Bowman}]{rein2024gpqa}
David Rein, Betty~Li Hou, Asa~Cooper Stickland, Jackson Petty, Richard~Yuanzhe Pang, Julien Dirani, Julian Michael, and Samuel~R Bowman. 2024.
\newblock Gpqa: A graduate-level google-proof q\&a benchmark.
\newblock In \emph{First Conference on Language Modeling}.

\bibitem[{Satpute et~al.(2024)Satpute, Giessing, Greiner-Petter, Schubotz, Teschke, Aizawa, and Gipp}]{satpute2024llms}
Ankit Satpute, Noah Giessing, Andre Greiner-Petter, Moritz Schubotz, Olaf Teschke, Akiko Aizawa, and Bela Gipp. 2024.
\newblock \href {https://arxiv.org/abs/2404.00344} {Can llms master math? investigating large language models on math stack exchange}.
\newblock \emph{arXiv preprint arXiv:2404.00344}.

\bibitem[{Shah et~al.(2024)Shah, Gandhi, Shah, Shah, Patil, and Bhowmick}]{shah2024infectious}
Chaitya Shah, Kashish Gandhi, Javal Shah, Kreena Shah, Nilesh Patil, and Kiran Bhowmick. 2024.
\newblock Infectious disease forecasting in india using llm's and deep learning.
\newblock \emph{arXiv preprint arXiv:2410.20168}.

\bibitem[{Sternberg(1997)}]{sternberg1997triarchic}
Robert~J Sternberg. 1997.
\newblock The triarchic theory of intelligence.
\newblock \emph{The Guilford Press}.

\bibitem[{Sujau et~al.(2025)Sujau, Wada, Vallée, Hillis, and Sušnjak}]{sujau2025disease}
Masood Sujau, Masako Wada, Emilie Vallée, Natalie Hillis, and Teo Sušnjak. 2025.
\newblock \href {https://doi.org/10.3390/make7020028} {Accelerating disease model parameter extraction: An llm-based ranking approach to select initial studies for literature review automation}.
\newblock \emph{Machine Learning and Knowledge Extraction}, 7(2):28.

\bibitem[{Sun et~al.(2025)Sun, Min, Chen, Zhao, Liu, Wang, Fang, and Wen}]{sun2025challenging}
Haoxiang Sun, Yingqian Min, Zhipeng Chen, Wayne~Xin Zhao, Zheng Liu, Zhongyuan Wang, Lei Fang, and Ji-Rong Wen. 2025.
\newblock Challenging the boundaries of reasoning: An olympiad-level math benchmark for large language models.
\newblock \emph{arXiv preprint arXiv:2503.21380}.

\bibitem[{Team et~al.(2023)Team, Anil, Borgeaud, Alayrac, Yu, Soricut, Schalkwyk, Dai, Hauth, Millican et~al.}]{team2023gemini}
Gemini Team, Rohan Anil, Sebastian Borgeaud, Jean-Baptiste Alayrac, Jiahui Yu, Radu Soricut, Johan Schalkwyk, Andrew~M Dai, Anja Hauth, Katie Millican, et~al. 2023.
\newblock Gemini: a family of highly capable multimodal models.
\newblock \emph{arXiv preprint arXiv:2312.11805}.

\bibitem[{Team(2024{\natexlab{a}})}]{2024qwen2.5}
Qwen Team. 2024{\natexlab{a}}.
\newblock \href {https://qwenlm.github.io/blog/qwen2.5/} {Qwen2.5: A party of foundation models}.

\bibitem[{Team(2024{\natexlab{b}})}]{qwen2024qwq}
Qwen Team. 2024{\natexlab{b}}.
\newblock \href {https://qwenlm.github.io/blog/qwq-32b-preview/} {Qwq: Reflect deeply on the boundaries of the unknown}.

\bibitem[{Wang et~al.(2024{\natexlab{a}})Wang, Wang, Xue, Xia, Cao, Liu, Pan, and Wong}]{wang2024appbench}
Hongru Wang, Rui Wang, Boyang Xue, Heming Xia, Jingtao Cao, Zeming Liu, Jeff Pan, and Kam-Fai Wong. 2024{\natexlab{a}}.
\newblock Appbench: Planning of multiple apis from various apps for complex user instruction.
\newblock In \emph{Proceedings of the 2024 Conference on Empirical Methods in Natural Language Processing}, pages 15322--15336.

\bibitem[{Wang et~al.(2024{\natexlab{b}})Wang, Wang, Su, Tong, and Song}]{wang2024rethinking}
Qineng Wang, Zihao Wang, Ying Su, Hanghang Tong, and Yangqiu Song. 2024{\natexlab{b}}.
\newblock Rethinking the bounds of llm reasoning: Are multi-agent discussions the key?
\newblock \emph{arXiv preprint arXiv:2402.18272}.

\bibitem[{Wu et~al.(2023)Wu, Bansal, Zhang, Wu, Li, Zhu, Jiang, Zhang, Zhang, Liu, Awadallah, White, Burger, and Wang}]{wu2023autogen}
Qingyun Wu, Gagan Bansal, Jieyu Zhang, Yiran Wu, Beibin Li, Erkang Zhu, Li~Jiang, Xiaoyun Zhang, Shaokun Zhang, Jiale Liu, Ahmed~Hassan Awadallah, Ryen~W White, Doug Burger, and Chi Wang. 2023.
\newblock \href {https://arxiv.org/abs/2308.08155} {Autogen: Enabling next-gen llm applications via multi-agent conversation}.
\newblock \emph{Preprint}, arXiv:2308.08155.

\bibitem[{Xie et~al.(2024)Xie, Zhang, Chen, Zhu, Lou, Tian, Xiao, and Su}]{xie2024travelplanner}
Jian Xie, Kai Zhang, Jiangjie Chen, Tinghui Zhu, Renze Lou, Yuandong Tian, Yanghua Xiao, and Yu~Su. 2024.
\newblock Travelplanner: A benchmark for real-world planning with language agents.
\newblock \emph{arXiv preprint arXiv:2402.01622}.

\bibitem[{Yang et~al.(2025)Yang, Chen, Zhang, Zhao, Qian, Wang, Wang, Koripella, Movahedi, Li et~al.}]{yang2025embodiedbench}
Rui Yang, Hanyang Chen, Junyu Zhang, Mark Zhao, Cheng Qian, Kangrui Wang, Qineng Wang, Teja~Venkat Koripella, Marziyeh Movahedi, Manling Li, et~al. 2025.
\newblock Embodiedbench: Comprehensive benchmarking multi-modal large language models for vision-driven embodied agents.
\newblock \emph{arXiv preprint arXiv:2502.09560}.

\bibitem[{Yang et~al.(2018)Yang, Qi, Zhang, Bengio, Cohen, Salakhutdinov, and Manning}]{yang2018hotpotqa}
Zhilin Yang, Peng Qi, Saizheng Zhang, Yoshua Bengio, William~W Cohen, Ruslan Salakhutdinov, and Christopher~D Manning. 2018.
\newblock Hotpotqa: A dataset for diverse, explainable multi-hop question answering.
\newblock \emph{arXiv preprint arXiv:1809.09600}.

\bibitem[{Yao et~al.(2022)Yao, Chen, Yang, and Narasimhan}]{yao2022webshop}
Shunyu Yao, Howard Chen, John Yang, and Karthik Narasimhan. 2022.
\newblock Webshop: Towards scalable real-world web interaction with grounded language agents.
\newblock \emph{Advances in Neural Information Processing Systems}, 35:20744--20757.

\bibitem[{Yao et~al.(2024)Yao, Shinn, Razavi, and Narasimhan}]{yao2024tau}
Shunyu Yao, Noah Shinn, Pedram Razavi, and Karthik Narasimhan. 2024.
\newblock tau-bench: A benchmark for tool-agent-user interaction in real-world domains.
\newblock \emph{arXiv preprint arXiv:2406.12045}.

\bibitem[{Zhang et~al.(2025)Zhang, Li, Qian, Liu, Yu, Han, Fung, McKeown, Zhai, Li et~al.}]{zhang2025law}
Yuji Zhang, Sha Li, Cheng Qian, Jiateng Liu, Pengfei Yu, Chi Han, Yi~R Fung, Kathleen McKeown, Chengxiang Zhai, Manling Li, et~al. 2025.
\newblock The law of knowledge overshadowing: Towards understanding, predicting, and preventing llm hallucination.
\newblock \emph{arXiv preprint arXiv:2502.16143}.

\bibitem[{Zhang et~al.(2024)Zhang, Bo, Ma, Li, Chen, Dai, Zhu, Dong, and Wen}]{zhang2024survey}
Zeyu Zhang, Xiaohe Bo, Chen Ma, Rui Li, Xu~Chen, Quanyu Dai, Jieming Zhu, Zhenhua Dong, and Ji-Rong Wen. 2024.
\newblock A survey on the memory mechanism of large language model based agents.
\newblock \emph{arXiv preprint arXiv:2404.13501}.

\bibitem[{Zhou et~al.(2023)Zhou, Xu, Zhu, Zhou, Lo, Sridhar, Cheng, Ou, Bisk, Fried, Alon, and Neubig}]{zhou2023webarena}
Shuyan Zhou, Frank Xu, Hao Zhu, Xuhui Zhou, Robert Lo, Abishek Sridhar, Xianyi Cheng, Tianyue Ou, Yonatan Bisk, Daniel Fried, Uri Alon, and Graham Neubig. 2023.
\newblock \href {https://openreview.net/forum?id=zlsj9akpaa} {Webarena: A realistic web environment for building autonomous agents}.
\newblock In \emph{NeurIPS 2023 Foundation Models for Decision Making Workshop}.

\bibitem[{Zhu et~al.(2025)Zhu, Du, Hong, Yang, Guo, Wang, Wang, Qian, Tang, Ji et~al.}]{zhu2025multiagentbench}
Kunlun Zhu, Hongyi Du, Zhaochen Hong, Xiaocheng Yang, Shuyi Guo, Zhe Wang, Zhenhailong Wang, Cheng Qian, Xiangru Tang, Heng Ji, et~al. 2025.
\newblock Multiagentbench: Evaluating the collaboration and competition of llm agents.
\newblock \emph{arXiv preprint arXiv:2503.01935}.

\end{thebibliography}

\clearpage
\appendix

\section*{Appendix}
\label{sec:appendix}

\begin{figure*}[t]
\centering
\small
\resizebox{1.0\textwidth}{!}{
\begin{tcolorbox}[colback=gray!5!white, colframe=blue!75!black, 
title=System Prompt for Problem Difficulty Categorization, boxrule=0.3mm, width=\textwidth, arc=3mm, auto outer arc=true]
\#\#\# \textbf{Task}\\
You are a helpful assistant to help me decide the difficulty of a math modeling problem. You are given the problem and their according images, pdf, and data (if exists). You should decide the difficulty through the following aspects.\\
\\
\#\#\# \textbf{Evaluation}\\
1. \textbf{Data Accessibility}: The math modeling problem I provide to you needs real-world data to solve and experiments to validate your model (data here is not just a few numbers, but large-scale data for validation purpose). Thus, for one problem, it may need data from diverse aspect. Based on the problem I provide, you should choose from the following options:\\
A. The data is provided in the problem, and using this data is enough to solve the problem.\\
B. The data is not provided in the problem, but based on the problem, it's easy to search the large amount of related data online.\\
C. The data is not provided in the problem, and it's very hard to search large amount of related data online because of the complexity or expertise required.\\
\\
2. \textbf{Modeling Difficulty}: The modeling difficulty is based on the complexity of the problem itself. Given the question, whether you are able to think of at least famous mathematical models that could be applied for analysis. You should choose from the following options:\\
A. You can think about more than 3 famous mathematical models (all should be about established methods) that could be applied for analysis.\\
B. You can come up with your own modeling ideas (not all of them are established methods, but they inspire your own model) that could be applied.\\
C. It's very hard to come up with any mathematical models that could be applied for analysis at first look, seems no established model could be adapted or applied for analysis.\\
\\
3. \textbf{Image Clarity}: The clarity of the image provided in the problem. You should choose from the following options:\\
A. The problem does not contain any images, or the image is only for illustrative purpose and not necessary for solving the problem.\\
B. The image contains important information for problem solving (include data, etc.), and I can clearly read it from the image captions and can use them for analysis and problem solving.\\
C. The image contains important information for problem solving (include data, etc.), but it's hard for you to get these information directly from the text or image captions, and without these information the problem becomes arbitrary or hard to solve.\\
\\
For each of the three aspects, you should choose the most appropriate option based on the problem I provide.
\end{tcolorbox}
}
\caption{The instruction used for categorizing the problem's difficulty.}
\label{prompt:problem_categorize}
\end{figure*}

\section{Data Curation Details}
\label{apdx:data_curation}
We include the system prompt used to instruct the model for categorization in \Cref{prompt:problem_categorize}. Following this, we conduct a thorough manual quality check of all modeling questions, with particular attention to those that received at least one ``C'' rating in the automatic categorization. For these cases, we either discard the questions or revise them to be simpler and more LLM-friendly. Additionally, we perform rigorous accessibility and feasibility checks on all data to ensure that each problem is appropriate for text-only LLMs.

We categorize the difficulty levels based on a heuristic derived from prior rating schemes. Specifically, problems receiving three ``A'' ratings are classified as \textit{easy}, those with exactly one rating lower than ``A'' are considered \textit{medium}, and the remaining problems are categorized as \textit{hard}. This classification underpins the difficulty distribution reported in \Cref{tab:statistics}.

\section{Math Modeling Reference Method Details}
\label{apdx:reference_model}
In the idea proposal phase, we instruct the model to generate multiple viable modeling approaches by referencing established mathematical modeling methodologies. These references serve as high-level conceptual guides, but typically require further adaptation to be applicable in grounded, real-world scenarios. Please refer to \Cref{fig:modeling_reference} for these common modeling methods.

\begin{figure*}[t]
\centering
\small
\resizebox{1.0\textwidth}{!}{
\begin{tcolorbox}[colback=gray!5!white, colframe=blue!75!black,
title=Systematic Overview of Modeling Approaches, boxrule=0.3mm, width=\textwidth, arc=3mm, auto outer arc=true]
\#\#\# \textbf{Model Categories and Techniques}\\
\\
\textbf{1. Evaluation Models (Decision-Making \& Multi-Criteria Analysis)}\\
- \textbf{Analytic Hierarchy Process (AHP)}: Used for ranking and decision-making based on pairwise comparisons.\\
- \textbf{Grey Relational Analysis (GRA)}: Evaluates relationships between different factors with incomplete or uncertain data.\\
- \textbf{Fuzzy Comprehensive Evaluation}: Applies fuzzy logic to assess multi-criteria problems.\\
- \textbf{Technique for Order Preference by Similarity to Ideal Solution (TOPSIS)}: Ranks alternatives based on their distance to an ideal solution.\\
- \textbf{Data Envelopment Analysis (DEA)}: Measures the efficiency of decision-making units.\\
- \textbf{Composite Evaluation Methods}: Combines multiple evaluation techniques for a comprehensive decision model.\\
\\
\textbf{2. Prediction Models (Forecasting \& Time-Series Analysis)}\\
- \textbf{Regression Analysis Prediction}: Uses statistical relationships between variables to make predictions.\\
- \textbf{Time Series Models}: Forecasting techniques based on past data trends (e.g., ARIMA).\\
- \textbf{Grey Prediction Model (GM)}: Works well for small-sample forecasting with limited data.\\
- \textbf{Markov Chain Prediction}: Uses probability transitions for state-based predictions.\\
- \textbf{Artificial Neural Networks (ANN)}: Machine learning models for complex pattern recognition.\\
- \textbf{Support Vector Machines (SVM)}: Effective for high-dimensional predictive modeling.\\
\\
\textbf{3. Classification Models (Machine Learning \& Supervised Learning)}\\
- \textbf{Logistic Regression}: Used for binary classification problems.\\
- \textbf{Decision Tree}: A rule-based classification model.\\
- \textbf{Random Forest}: An ensemble of decision trees for better accuracy.\\
- \textbf{Naive Bayes (Bayesian Classification)}: Based on probability and Bayes’ theorem.\\
- \textbf{K-Nearest Neighbors (KNN)}: Classifies based on the majority vote of nearest data points.\\
\\
\textbf{4. Statistical Analysis Models (Hypothesis Testing \& Data Analysis)}\\
- \textbf{t-Test}: Compares means between two groups.\\
- \textbf{Analysis of Variance (ANOVA)}: Tests differences among multiple groups.\\
- \textbf{Chi-Square Test}: Analyzes categorical data for independence.\\
- \textbf{Correlation Analysis}: Measures relationships between variables.\\
- \textbf{Regression Analysis}: Determines dependencies between variables.\\
- \textbf{Logistic Regression}: Predicts probability outcomes (also used in classification).\\
- \textbf{Cluster Analysis}: Groups similar data points together.\\
- \textbf{Principal Component Analysis (PCA)}: Reduces dimensionality while preserving variance.\\
- \textbf{Factor Analysis}: Identifies underlying relationships between variables.\\
\end{tcolorbox}
}
\caption{Common modeling approaches used by the Idea Proposer as references.}
\label{fig:modeling_reference}
\end{figure*}

Interestingly, we observe that our ModelingAgent not only adapts these methods effectively but also demonstrates the ability to synthesize different techniques and propose novel approaches not present in the original references. This highlights the model's capacity for innovative and creative problem-solving beyond mere replication.

\section{ModelingAgent Details}
\label{apdx:modeling_agent}

We employ the critic module in multiple stages of our multi-agent framework, applying distinct rubrics tailored to different evaluation purposes. These rubrics are derived from the official judge commentaries released for each question, where we manually summarize the core aspects emphasized by judges into focused criteria to better guide the modeling refinement process.

Specifically, we design the following rubrics for each aspect where the critic module is applied:

\noindent \textbf{1. Critics for Modeling Idea Proposing (Idea Proposer):}
\begin{itemize}[topsep=2pt, partopsep=-5pt, leftmargin=8pt, itemsep=-2pt]
    \item \textbf{Relevance:} Determine if the proposed approach adequately addresses the subtask objective, and identify any gaps or potential improvements.
    \item \textbf{Mathematical Rigor:} Evaluate whether the proposed idea is mathematically sound and accounts for all critical factors, highlighting missing components and suggesting refinements.
    \item \textbf{Practical Feasibility:} Assess whether the proposed idea is realistically feasible given limited online resources, basic computational tools (such as Python libraries), and data accessibility, identifying potential challenges.
\end{itemize}

\noindent \textbf{2. Critics for Mathematical Formulation (Modeling Implementor):}
\begin{itemize}[topsep=2pt, partopsep=-5pt, leftmargin=8pt, itemsep=-2pt]
    \item \textbf{Comprehensiveness:} Assess whether the mathematical formulation thoroughly addresses the subtask objective, and identify any missing elements or areas for refinement.
    \item \textbf{Mathematical Rigor:} Evaluate if the formulation is mathematically sound, employing formalized expressions and highlighting any gaps or inconsistencies.
    \item \textbf{Practical Feasibility:} Determine whether the formulation is realistically executable with limited computational resources and accessible data, noting any implementation challenges.
\end{itemize}

\noindent \textbf{3. Critics for Data Searching (Data Searcher):}
\begin{itemize}[topsep=2pt, partopsep=-5pt, leftmargin=8pt, itemsep=-2pt]
    \item \textbf{Data Quality:} Examine whether the collected data is relevant, accurate, sufficient, and properly organized.
    \item \textbf{Data Reliability:} Assess the trustworthiness of the data based on source credibility, consistency, and potential biases.
    \item \textbf{File Structure Completeness:} Verify whether the required CSV and MD files have been correctly created with appropriate content and structure.
\end{itemize}

\noindent \textbf{4. Critics for Modeling Implementation and Analysis (Modeling Implementor):}
\begin{itemize}[topsep=2pt, partopsep=-5pt, leftmargin=8pt, itemsep=-2pt]
    \item \textbf{Model Approach:} Check if the modeling approach addresses all critical factors with justified assumptions and includes quantitative sensitivity analysis.
    \item \textbf{Model Implementation:} Assess whether the code is clean, modular, efficient, reproducible, and properly tested.
    \item \textbf{Report Quality:} Verify that the report is professional, follows the template, and includes clear, well-labeled figures with proper interpretation.
\end{itemize}

\section{ModelingJudge Details}
\label{apdx:modeling_judge}

We construct the ModelingJudge framework by defining multiple expert roles and evaluating solution quality across three groundedness aspects and one innovativeness aspect. To ensure that judgments from each expert role are clear, consistent, and well-founded, we carefully design detailed rubrics for each evaluation aspect, along with corresponding scoring scales. The detailed instructions are presented in \Cref{prompt:judge_analysis} to \Cref{prompt:judge_innovation}. For each judging aspect, the final score is computed by averaging the rubric scores, and the overall evaluation is then obtained by averaging across all expert roles.

\section{Human Evaluation}
\label{apdx:human_eval}

To assess the quality of math modeling reports generated by various models, we conducted a comprehensive human evaluation. Recognizing the inherent subjectivity of human judgments, we adopted an arena-style evaluation framework, inspired by methodologies such as the Chatbot Arena. This approach enables direct comparisons between model outputs, allowing evaluators to rank responses based on perceived quality.

Specifically, we randomly selected modeling questions covering topics from the Mathematical Contest in Modeling (MCM) and the Interdisciplinary Contest in Modeling (ICM), spanning difficulty levels from high school to undergraduate. The evaluation was conducted under three distinct settings. In the first setting (a), for each selected question, we collected math modeling implementations generated by seven different models, all using the same \textit{ModelingAgent} framework to ensure fairness. Participants were then asked to rank the top three solutions without knowing which model produced each one. In the second setting (b), for the same questions, we compared solutions generated using different modeling methods, including a top human performance upperbound. To maintain consistency, all model-generated solutions in this setting were produced by GPT-4o. Participants were again asked to rank the top three solutions. In the final setting (c), participants were shown the four solutions as above from different methods, and asked to identify which one was most likely authored by a human expert team. These settings correspond to the three sub-graphs in \Cref{fig:human_evaluation}.

We recruited 12 volunteer participants to serve as human evaluators, 60\% of whom had prior experience in national or international mathematical modeling competitions. All participants had academic backgrounds in computer science or mathematics, ranging from undergraduate to postgraduate levels. The evaluation took approximately 10 minutes to complete. No financial compensation was provided, as participation was entirely voluntary and motivated by genuine interest in the topic and the intellectual challenge. All participants provided informed consent for the use of their evaluation data in this study. The data collection process underwent ethical review and involved no sensitive or personally identifiable information, ensuring full compliance with ethical research standards.

\section{Top Human Performance}
\label{apdx:top_human}
\begin{figure*}
    \centering
    \includegraphics[width=\linewidth]{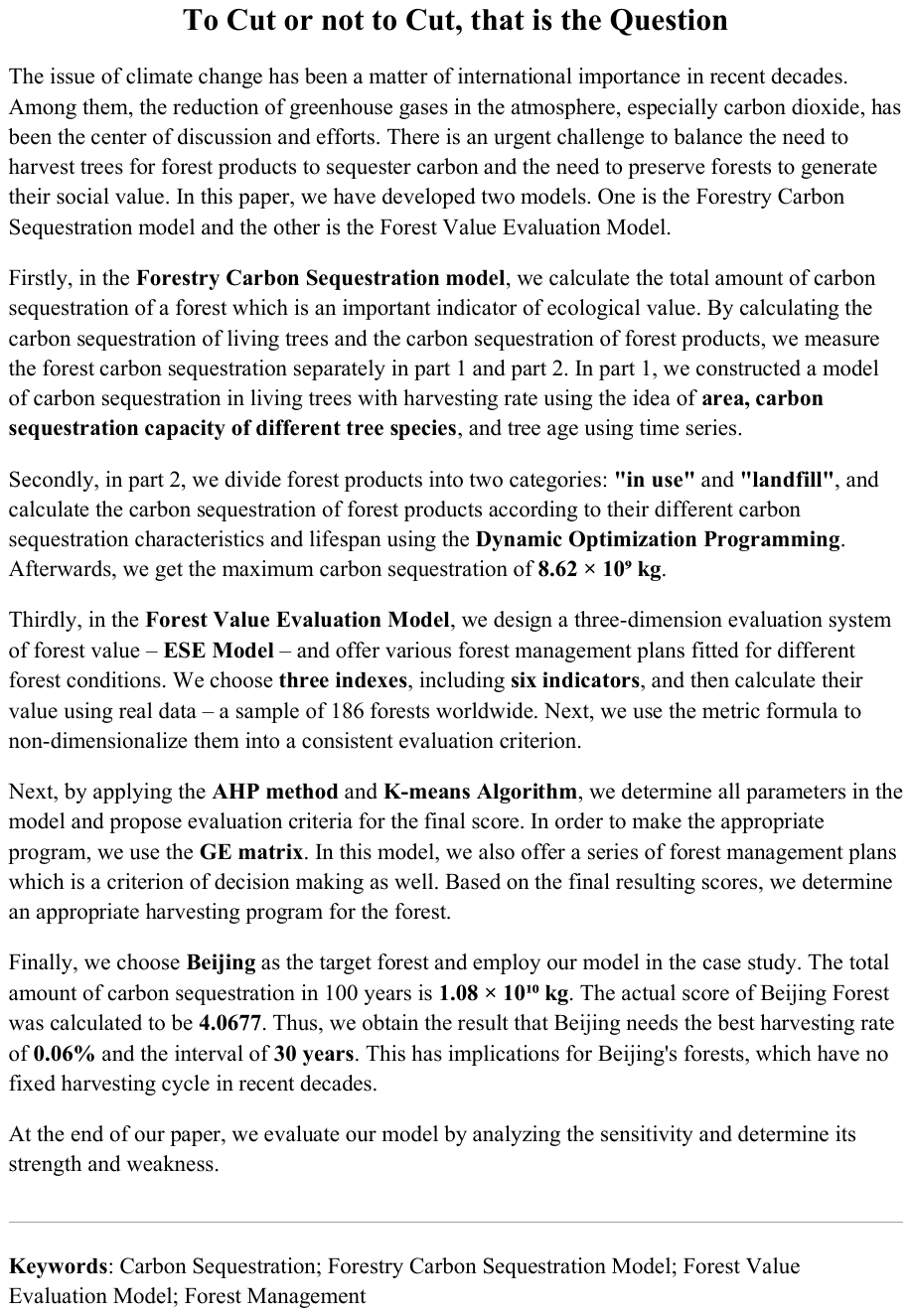}
    \caption{The summary sheet of top-human performance that wins the highest award.}
    \label{fig:summary_sheet}
\end{figure*}

In this section, we present a top human solution that achieved the highest award in the MCM competition, corresponding to the modeling problem illustrated in \Cref{fig:skill}. The summary sheet of this award-winning solution is provided in \Cref{fig:summary_sheet}.

To facilitate a side-by-side comparison, we also showcase two complete modeling implementations: one produced by a top human team from the award-winning report and the other generated by the ModelingAgent. These are presented in \Cref{appendix:human_solution_full} and \Cref{appendix:model_solution_full}, respectively. This example also serves as an example for the human evaluation discussed in \Cref{sec:exp_analysis}.

\section{Critic Performance}
\label{apdx:critic_performance}

We additionally present two examples of critic performance in \Cref{fig:critic_performance_new1} and \Cref{fig:critic_performance_new2}. The first example illustrates the Data Searcher, highlighting the initial critic feedback and how the data search process improves in the second round. The second example demonstrates how the critic thoroughly evaluates a modeling idea generated by Idea Proposer, showcasing the changes in the critic’s feedback and scores after the agent refines its approach.

\section{Prompt Instruction Details}
\label{apdx:prompt_details}

This section provides detailed descriptions of the prompts used in our main experiments, as shown in \Cref{prompt:baseagent} to \Cref{prompt:modelingagent_report_writer}. These prompts cover the Vanilla Generation baseline, the Tool Agent, and our proposed ModelingAgent system. Note that within our agent system, the corresponding prompts are also applied individually to each sub-agent.

\begin{figure*}[t]
\centering
\small
\resizebox{0.9\textwidth}{!}{
\begin{tcolorbox}[colback=gray!5!white, colframe=brown!75!black, 
title=System Prompt for ModelingJudge Groundedness of Analysis, boxrule=0.3mm, width=\textwidth, arc=3mm, auto outer arc=true]
\{\{Expert Role Description\}\}\\
\\
You are currently evaluating mathematical modeling papers. Your task is to assess how well the solution's analysis is grounded in mathematical and scientific principles. You should evaluate based on the role you are given. Score each aspect from 0-1, starting at 0 and requiring justification for any increase:\\
\\
1. \textbf{Analytical Depth (0-1):}\\
   0.00: No meaningful analysis\\
        Example: Superficial observations without reasoning\\
   0.25: Basic analysis\\
        Example: Simple descriptive analysis without connections\\
   0.50: Standard analysis\\
        Example: Clear reasoning with some depth\\
   0.75: Advanced analysis\\
        Example: Deep insights with strong connections\\
   1.00: Exceptional analysis\\
        Example: Novel insights with comprehensive reasoning\\
\\
2. \textbf{Mathematical Rigor (0-1):}\\
   0.00: No mathematical support\\
        Example: Claims without mathematical backing\\
   0.25: Basic mathematics\\
        Example: Simple calculations without justification\\
   0.50: Standard rigor\\
        Example: Clear mathematical reasoning\\
   0.75: Strong rigor\\
        Example: Detailed proofs and derivations\\
   1.00: Exceptional rigor\\
        Example: Complete mathematical framework\\
\\
3. \textbf{Results Interpretation (0-1):}\\
   0.00: No interpretation\\
        Example: Raw results without context\\
   0.25: Basic interpretation\\
        Example: Simple description of results\\
   0.50: Clear interpretation\\
        Example: Results explained with context\\
   0.75: Thorough interpretation\\
        Example: Deep analysis of implications\\
   1.00: Exceptional interpretation\\
        Example: Comprehensive analysis with insights\\
\\
4. \textbf{Critical Analysis (0-1):}\\
   0.00: No critical thinking\\
        Example: Accepts all results without question\\
   0.25: Basic criticism\\
        Example: Notes obvious limitations\\
   0.50: Standard analysis\\
        Example: Identifies key strengths/weaknesses\\
   0.75: Strong analysis\\
        Example: Deep examination of assumptions\\
   1.00: Exceptional analysis\\
        Example: Comprehensive critique with alternatives\\
\\
5. \textbf{Future Implications (0-1):}\\
   0.00: No discussion\\
        Example: Ends at results\\
   0.25: Basic implications\\
        Example: Simple next steps\\
   0.50: Clear implications\\
        Example: Reasonable future directions\\
   0.75: Strong implications\\
        Example: Detailed future research paths\\
   1.00: Exceptional vision\\
        Example: Novel research directions with justification\\
\\
Critical Scoring Rules:\\
- Start at 0 for each aspect\\
- Must justify ANY increase in score with specific evidence\\
- No mathematical justification caps score at 0.25\\
- Missing critical analysis caps score at 0.50\\
- Most solutions should score between 0.25-0.50\\
- Perfect scores (1.00) should be extremely rare\\
\end{tcolorbox}
}
\caption{The instruction for expert model to evaluate the solution's groundedness of analysis.}
\label{prompt:judge_analysis}
\end{figure*}

\begin{figure*}[t]
\centering
\small
\resizebox{0.9\textwidth}{!}{
\begin{tcolorbox}[colback=gray!5!white, colframe=brown!75!black, 
title=System Prompt for ModelingJudge Groundedness of Data, boxrule=0.3mm, width=\textwidth, arc=3mm, auto outer arc=true]
\{\{Expert Role Description\}\}\\
\\
You are currently evaluating mathematical modeling papers. Your task is to assess how well the solution is grounded in data and evidence. You should evaluate based on the role you are given. Score each aspect from 0-1, starting at 0 and requiring justification for any increase:\\
\\
1. \textbf{Data Quality (0-1):}\\
   0.00: No data or invalid data\\
        Example: Made-up numbers without sources\\
   0.25: Poor quality/unreliable\\
        Example: Single unreliable source, outdated data\\
   0.50: Acceptable but limited\\
        Example: Reliable source but incomplete dataset\\
   0.75: Good with minor issues\\
        Example: Multiple reliable sources, small gaps\\
   1.00: Excellent data quality\\
        Example: Multiple verified sources, comprehensive coverage\\
\\
2. \textbf{Data Processing (0-1):}\\
   0.00: No processing/invalid\\
        Example: Raw data used without cleaning\\
   0.25: Basic processing only\\
        Example: Simple averaging without outlier removal\\
   0.50: Standard processing\\
        Example: Basic cleaning and normalization\\
   0.75: Advanced processing\\
        Example: Sophisticated cleaning with justification\\
   1.00: Comprehensive processing\\
        Example: Full pipeline with validation at each step\\
\\
3. \textbf{Statistical Analysis (0-1):}\\
   0.00: No analysis/incorrect\\
        Example: No statistical methods used\\
   0.25: Basic statistics only\\
        Example: Mean/median without confidence intervals\\
   0.50: Standard analysis\\
        Example: Basic hypothesis testing\\
   0.75: Advanced analysis\\
        Example: Multiple statistical methods with validation\\
   1.00: Rigorous analysis\\
        Example: Comprehensive statistical framework with robustness checks\\
\\
4. \textbf{Data Integration (0-1):}\\
   0.00: No integration\\
        Example: Data disconnected from model\\
   0.25: Poor integration\\
        Example: Forced fit without justification\\
   0.50: Partial integration\\
        Example: Some aspects well-integrated, others not\\
   0.75: Good integration\\
        Example: Most data well-integrated with clear reasoning\\
   1.00: Perfect integration\\
        Example: All data seamlessly integrated with full justification\\
\\
5. \textbf{Validation} \textbf{\&} \textbf{Testing (0-1):}\\
   0.00: No validation\\
        Example: Results accepted without testing\\
   0.25: Minimal testing\\
        Example: Basic sanity checks only\\
   0.50: Standard validation\\
        Example: Cross-validation without sensitivity analysis\\
   0.75: Thorough validation\\
        Example: Multiple validation methods\\
   1.00: Comprehensive validation\\
        Example: Full validation suite with sensitivity analysis\\
\\
Critical Scoring Rules:\\
- Start at 0 for each aspect\\
- Must justify ANY increase in score with specific evidence\\
- Missing data source documentation caps score at 0.25\\
- No data processing caps score at 0.25\\
- No validation caps overall score at 0.50\\
- Most solutions should score between 0.25-0.50\\
\end{tcolorbox}
}
\caption{The instruction for expert model to evaluate the solution's groundedness of data.}
\label{prompt:judge_data}
\end{figure*}

\begin{figure*}[t]
\centering
\small
\resizebox{0.9\textwidth}{!}{
\begin{tcolorbox}[colback=gray!5!white, colframe=brown!75!black, 
title=System Prompt for ModelingJudge Groundedness of Modeling, boxrule=0.3mm, width=\textwidth, arc=3mm, auto outer arc=true]
\{\{Expert Role Description\}\}\\
\\
You are currently evaluating mathematical modeling papers. Your task is to assess how well the solution's modeling approach is grounded in mathematical and scientific principles. You should evaluate based on the role you are given. Score each aspect from 0-1, starting at 0 and requiring justification for any increase:\\
\\
1. \textbf{Mathematical Foundation (0-1):}\\
   0.00: Fundamentally flawed or missing\\
        Example: No equations, incorrect mathematical concepts\\
   0.25: Basic but problematic\\
        Example: Simple equations without proper variables defined\\
   0.50: Sound but incomplete\\
        Example: Correct equations but missing key relationships\\
   0.75: Strong with minor gaps\\
        Example: Well-formulated with some assumptions not fully justified\\
   1.00: Excellent and rigorous\\
        Example: Complete mathematical framework with all relationships justified\\
\\
2. \textbf{Real-World Integration (0-1):}\\
   0.00: No connection to reality\\
        Example: Pure abstract model without practical context\\
   0.25: Superficial consideration\\
        Example: Mentioning real factors without incorporating them\\
   0.50: Partial integration\\
        Example: Some key factors included but others missing\\
   0.75: Good but not comprehensive\\
        Example: Most factors included but some interactions overlooked\\
   1.00: Complete integration\\
        Example: All relevant factors and interactions properly modeled\\
\\
3. \textbf{Technical Sophistication (0-1):}\\
   0.00: Elementary/inappropriate\\
        Example: Using linear regression for clearly nonlinear problems\\
   0.25: Basic techniques only\\
        Example: Simple statistical methods without justification\\
   0.50: Appropriate but limited\\
        Example: Correct methods but not fully exploited\\
   0.75: Advanced with minor issues\\
        Example: Sophisticated methods with some gaps in implementation\\
   1.00: State-of-the-art\\
        Example: Cutting-edge techniques properly implemented\\
\\
4. \textbf{Validation Approach (0-1):}\\
   0.00: No validation\\
        Example: Results presented without any verification\\
   0.25: Minimal testing\\
        Example: Basic sanity checks only\\
   0.50: Partial validation\\
        Example: Some test cases but not comprehensive\\
   0.75: Thorough but not complete\\
        Example: Multiple validation methods but missing edge cases\\
   1.00: Comprehensive validation\\
        Example: Multiple methods, edge cases, sensitivity analysis\\
\\
5. \textbf{Implementation Quality (0-1):}\\
   0.00: Poor/incorrect\\
        Example: Errors in implementation, wrong formulas\\
   0.25: Basic but flawed\\
        Example: Correct concept but significant implementation errors\\
   0.50: Workable but needs improvement\\
        Example: Functions correctly but inefficient or unclear\\
   0.75: Good with minor issues\\
        Example: Well-implemented but some optimization possible\\
   1.00: Excellent implementation\\
        Example: Efficient, clear, and well-documented code\\
\\
Critical Scoring Rules:\\
- Start at 0 for each aspect\\
- Must justify ANY increase in score with specific evidence\\
- Missing any critical element caps score at 0.25\\
- Lack of validation caps overall score at 0.50\\
- Surface-level treatment caps score at 0.25\\
- Most solutions should score between 0.25-0.50\\
\end{tcolorbox}
}
\caption{The instruction for expert model to evaluate the solution's groundedness of modeling.}
\label{prompt:judge_modeling}
\end{figure*}

\begin{figure*}[t]
\centering
\small
\resizebox{0.9\textwidth}{!}{
\begin{tcolorbox}[colback=gray!5!white, colframe=brown!75!black, 
title=System Prompt for ModelingJudge Innovativeness, boxrule=0.3mm, width=\textwidth, arc=3mm, auto outer arc=true]
\{\{Expert Role Description\}\}\\
\\
You are currently evaluating mathematical modeling papers. Your task is to assess the innovativeness and originality of the solution approach. You should evaluate based on the role you are given. Score each aspect from 0-1, starting at 0 and requiring justification for any increase:\\
\\
1. \textbf{Methodological Innovation (0-1):}\\
   0.00: Standard/textbook approach\\
        Example: Using basic linear regression without modification\\
   0.25: Minor adaptations\\
        Example: Small tweaks to existing methods\\
   0.50: Meaningful modifications\\
        Example: Significant adaptations to standard approaches\\
   0.75: Novel combinations\\
        Example: Creative synthesis of multiple methods\\
   1.00: Groundbreaking approach\\
        Example: Entirely new methodology with strong justification\\
\\
2. \textbf{Problem Framing (0-1):}\\
   0.00: Conventional perspective\\
        Example: Following typical problem formulation\\
   0.25: Slight reframing\\
        Example: Minor changes to standard approach\\
   0.50: Fresh perspective\\
        Example: New angle on known problem\\
   0.75: Novel framing\\
        Example: Unique problem decomposition\\
   1.00: Revolutionary perspective\\
        Example: Paradigm-shifting problem formulation\\
\\
3. \textbf{Solution Creativity (0-1):}\\
   0.00: Standard solution\\
        Example: Direct application of known methods\\
   0.25: Minor creativity\\
        Example: Small creative elements in standard approach\\
   0.50: Notable creativity\\
        Example: Original elements in key areas\\
   0.75: Significant creativity\\
        Example: Multiple creative components\\
   1.00: Exceptional creativity\\
        Example: Entirely novel solution approach\\
\\
4. \textbf{Technical Advancement (0-1):}\\
   0.00: No advancement\\
        Example: Uses only existing techniques\\
   0.25: Minor improvements\\
        Example: Small technical optimizations\\
   0.50: Meaningful advances\\
        Example: New technical contributions\\
   0.75: Significant advances\\
        Example: Multiple technical innovations\\
   1.00: Major breakthrough\\
        Example: Revolutionary technical approach\\
\\
5. \textbf{Impact Potential (0-1):}\\
   0.00: Minimal impact\\
        Example: No new insights or applications\\
   0.25: Limited impact\\
        Example: Minor improvements to existing methods\\
   0.50: Moderate impact\\
        Example: Useful new approach for specific cases\\
   0.75: High impact\\
        Example: Broadly applicable new methods\\
   1.00: Transformative\\
        Example: Could change the field significantly\\
\\
Critical Scoring Rules:\\
- Start at 0 for each aspect\\
- Must justify ANY increase in score with specific evidence\\
- Using only standard methods caps score at 0.25\\
- Lack of justification for novelty caps score at 0.50\\
- Most solutions should score between 0.25-0.50\\
- Perfect scores (1.00) should be extremely rare\\
- Innovation without proper validation caps at 0.25\\
\end{tcolorbox}
}
\caption{The instruction for expert model to evaluate the solution's innovativeness.}
\label{prompt:judge_innovation}
\end{figure*}

\begin{figure*}[htbp]
    \centering
    \scriptsize                
    \begin{tcolorbox}[
        title=\textbf{Data Searcher Critic Performance – Rounds 1 \& 2},
        colback=gray!5!white,
        colframe=black!60,
        fonttitle=\bfseries,
        left=2mm,right=2mm,top=1mm,bottom=1mm,
        boxrule=0.6pt,
        auto outer arc=true,
        width=\textwidth
    ]
\textbf{Modeling Question}

\begin{verbatim}
Honeybees are crucial for human survival due to their role in pollination and honey production, but their populations are declining 
due to factors such as viruses,pesticides, predators, habitat destruction, and environmental conditions. A model is needed to 
determine the population dynamics of a honeybee colony over time, taking into account factors like travel range, hive population, 
flower visits, and lifespan, which is influenced by activity level, pollen consumption, and protein abundance. The model 
should also predict the number of hives required to support the pollination of a specified area of land with crops.
\end{verbatim}

\rule{\linewidth}{0.4pt}

\textbf{Current Factor Under Investigation}

\begin{verbatim}
"variable": "External Threats",
"reason": "External threats, including predators, diseases, and pesticides, affect bee mortality and colony health. Understanding 
these threats is crucial for modeling their impact on population dynamics and developing mitigation strategies.",
"real_world_acquisition": "To find data on external threats, search for studies ordatasets that report on bee predators, 
disease prevalence, and pesticide exposure. Use search terms like 'honeybee predator impact', 'bee disease prevalence data', or 
'pesticide exposure on bees'. Research articles on pollinator health and environmental stressors can provide valuable 
information."
\end{verbatim}

\rule{\linewidth}{0.4pt}

\textbf{Round 1}

\begin{verbatim}
== Processing Critique for 'Hive Population' ==
{
  "scores": {
    "data_quality_score": 2,
    "reliability_score": 3,
    "file_structure_score": 2,
    "overall_score": 7
  },
  "weaknesses": [
    "Data is limited to chemical names, toxicity levels, and classes without broader context on bee mortality or colony health.",
    "Lacks comprehensiveness in covering other external threats such as predators and diseases.",
    "Data is not organized to support modeling of population dynamics.",
    "Lack of detailed documentation on how chemicals impact honeybee populations.",
    "No information on methodology or validation against empirical bee mortality data.",
    "File structure is inadequate with a PDF file instead of a structured CSV or MD file.",
    "Lack of documentation makes it difficult to assess data relevance and applicability."
  ],
  "recommendations": [
    "Broaden data scope to include other external threats such as predators and diseases.",
    "Improve documentation by creating a detailed MD file with data sources, processing steps, and limitations.",
    "Seek new data sources that provide a holistic view of external threats to honeybees, including empirical studies on bee mortality."
  ],
  "next_steps": [
    "Expand data collection to cover a wider range of external threats beyond chemical toxicity.",
    "Develop comprehensive documentation to enhance data transparency and reliability.",
    "Identify and integrate new data sources that offer a broader perspective on threats to honeybees.",
    "Seek alternative data sources with direct access to structured data formats.",
    "Use different tools or methods to handle zip files and ensure proper extraction.",
    "Once new data is sourced, create CSV and MD files to document and structure the data appropriately."
  ]
}
\end{verbatim}

\rule{\linewidth}{0.4pt}

\textbf{Round 2}

\begin{verbatim}
**Processed Feedback**:
{
  "scores": {
    "data_quality_score": 4,
    "reliability_score": 4,
    "file_structure_score": 5,
    "overall_score": 13
  },
  "weaknesses": [
    "Dataset is limited to primarily U.S.-based threats, which may not capture global conditions.",
    "Reliance on specific databases for pesticide toxicity data may introduce bias or gaps."
  ],
  "recommendations": [
    "Expand data scope to include information on external threats to honeybees in regions outside the United States.",
    "Integrate additional pesticide toxicity datasets to address potential gaps and improve comprehensiveness.",
    "Establish a process for regularly updating the dataset to incorporate new findings and emerging threats."
  ],
  "next_steps": [
    "Identify and incorporate additional data sources that provide global information on external threats to honeybees.",
    "Seek out and integrate more comprehensive pesticide toxicity datasets.",
    "Develop a regular update schedule for the dataset to ensure it remains current and relevant."
  ]
}
\end{verbatim}

    \end{tcolorbox}
    \caption{\textbf{Data Searcher Critic Performance.} Feedback from two review rounds and current factor search detail.}
    \label{fig:critic_performance_new1}
\end{figure*}

\begin{figure*}[htbp]
    \centering
    \scriptsize              
    \begin{tcolorbox}[
        title=\textbf{Idea Proposer Critic Performance – Rounds 1 \& 2},
        colback=gray!5!white,
        colframe=black!60,
        fonttitle=\bfseries,
        left=2mm,right=2mm,top=1mm,bottom=1mm,
        boxrule=0.6pt
    ]

\textbf{Modeling Question}

\begin{verbatim}
Honeybees are crucial for human survival due to their role in pollination and honey production, 
but their populations are declining due to factors such as viruses, pesticides, predators, 
habitat destruction, and environmental conditions. A model is needed to determine the population 
dynamics of a honeybee colony over time, taking into account factors like travel range, hive 
population, flower visits, and lifespan, which is influenced by activity level, pollen consumption, 
and protein abundance. The model should also predict the number of hives required to support the 
pollination of a specified area of land with crops.
\end{verbatim}

\rule{\linewidth}{0.4pt}

\textbf{Modeling Approach}

\begin{verbatim}
"modeling_approach": {
  "objective": "Determine the number of hives needed to effectively pollinate a specified area of land with crops.",
  "analysis": "The model should estimate the pollination capacity of a single hive based on factors such as the number 
  of foraging bees, their travel range, and the frequency of flower visits. It should also consider the specific 
  pollination needs of the crops in question, including flower density and pollination efficiency. The goal is to 
  calculate the total number of hives required to achieve optimal pollination coverage for the given area."
}
\end{verbatim}

\rule{\linewidth}{0.4pt}

\textbf{Round 1}

\begin{verbatim}
== Critique for Artificial Neural Networks Approach ==
{
  "scores": {
    "relevance_score": 5,
    "rigorous_score": 4,
    "practical_score": 3,
    "overall_score": 12
  },
  "relevance_feedback": "ANNs are highly relevant for capturing complex, non‑linear patterns in fish migration data. 
  They can integrate various factors, including temperature changes and biological behaviors, making them suitable for 
  this task.",
  "rigorous_feedback": "ANNs are mathematically robust and capable of modeling complex interactions. However, they 
  require large datasets for training and can be prone to overfitting. Regularization techniques and cross‑validation 
  are necessary to ensure model reliability.",
  "practical_feedback": "Implementing ANNs requires significant computational resources and expertise in machine learning.
  Access to comprehensive datasets is crucial, which might be a challenge. Tools like TensorFlow or PyTorch can facilitate 
  development."
}
\end{verbatim}

\rule{\linewidth}{0.4pt}

\textbf{Round 2}

\begin{verbatim}
== Critique for Ecological Niche Modeling Approach ==
{
  "scores": {
    "comprehensiveness_score": 5,
    "mathematical_rigor_score": 5,
    "practical_feasibility_score": 4,
    "overall_score": 14
  },
  "comprehensiveness_feedback": "The proposed model is comprehensive in its approach to predicting future habitats of 
  herring and mackerel. It effectively integrates ecological niche modeling with climate projections, considering both 
  environmental and biological factors. The model also addresses the economic impact on the fishing industry, providing 
  a holistic view of the problem. However, it could benefit from a more detailed explanation of how biological factors B 
  are quantified and incorporated into the habitat suitability index. Additionally, the model could explore other 
  environmental variables such as salinity and ocean currents, which may also influence fish distribution.",
  "mathematical_rigor_feedback": "The model is mathematically rigorous, using formal notation and logical derivations to 
  describe the relationships between temperature, habitat suitability, and population distribution. The equations are 
  well‑structured and align with ecological modeling practices. However, the model could improve by providing more 
  detailed methodologies for deriving functions f and g from historical data, ensuring that these functions accurately 
  reflect real‑world dynamics. Additionally, the integration of climate projections C could be more explicitly defined, 
  detailing how these projections are obtained and validated.",
  "practical_feasibility_feedback": "The model is practically feasible, leveraging existing tools like MaxEnt or BIOMOD 
  for ecological niche modeling. It is adaptable to future climate scenarios, making it relevant for real‑world 
  applications. However, the feasibility may be limited by the availability and quality of detailed environmental and 
  occurrence data, which are crucial for accurate predictions. The model's reliance on climate projections also 
  introduces uncertainty, as these projections can vary significantly. Addressing these data challenges and ensuring 
  robust validation of climate models would enhance practical feasibility."
}
\end{verbatim}

    \end{tcolorbox}
    \caption{\textbf{Idea Proposer Critic Performance.} Modeling question, approach, and feedback from two review rounds.}
    \label{fig:critic_performance_new2}
\end{figure*}

\begin{figure*}[t]
\centering
\small
\begin{tcolorbox}[colback=gray!5!white, colframe=blue!75!black, 
title=Human Answer, boxrule=0.3mm, width=\textwidth, arc=3mm, auto outer arc=true]
     
\textbf{Model Overview}

The model we have decided to implement is a filtration system and a ranking system.

First, we start from the inputs; budget, energy use, and schedules. These inputs will be fundamental to the filtration system, allowing for proper optimization. Those inputs would make their way into the Pre-Model section. This section collects and manages that data to allow for the model to work more smoothly. Then we move on to the Recharge (charge) rate, which would set up the model by providing all possible combinations. The combinations will be split up into 3 different sections for simplicity. Once we have all of the possible combinations, we can run them through filters. Through each filter, the possible combinations will decrease, giving us a more narrow answer.

- The first filter is the price filter, which relies on their budget; if the price exceeds the budget, then that combination is filtered out.

- The next filter is the usable capacity filter; for combinations that do not meet the storage needs, they would be filtered out.

Finally, running through those filters would not leave us with a definite answer, that’s why we implemented a ranking system. This ranking system will take several factors—average capacity, leftover price, and size—and output the top 3 ranked options, allowing the person to choose if they have another preference.

\rule{\linewidth}{0.2pt}

\textbf{Recharge Rate \& Number of Batteries  }

\#\#\#\# Scenario 1 – One Continuous \& One Instantaneous Battery
\[
R_{e_i}(N_c)_i = 4E_{c,\text{rate}},\qquad (N_c)_i=\left\lceil\frac{4E_{c,\text{rate}}}{R_{e_i}}\right\rceil
\]
\[
R_{e_i}(N_I)_i = 4E_{I,\text{rate}},\qquad (N_I)_i=\left\lceil\frac{4E_{I,\text{rate}}}{R_{e_i}}\right\rceil
\]
Price filter  
\[
\delta_1 = \Bigl\{ \tfrac{P_{B_i}}{P_{B_j}}\bigl[(N_c)_i(N_I)_j\bigr]\le\text{Budget} \Bigr\}
\]
Capacity filter  
\[
1<\frac{a_i(N_c)_i}{4E_c}<2,\qquad
1<\frac{a_i(N_I)_i}{4E_I}<2
\]

\#\#\# Scenario 2 – One Battery Type Covers Both Needs
\[
R_{e_i}N_i = 4E_{\text{tot,rate}},\qquad N_i=\left\lceil\frac{4E_{\text{tot,rate}}}{R_{e_i}}\right\rceil
\]
Price filter \(\delta_2 =\{P_{B_i}N_i\le\text{Budget}\}\)  
Capacity filter \(1<\tfrac{a_iN_i}{4E_{\text{tot}}}<2\)

\#\#\# Scenario 3 – Two Battery Types for Continuous/Instantaneous
\[
R_{e_i}(N_I)_i+R_{e_j}(N_I)_j = 4E_{I,\text{rate}},\quad
(N_I)_i=\Bigl\lceil\frac{4E_{I,\text{rate}}-R_{e_j}(N_I)_j}{R_{e_i}}\Bigr\rceil
\]
Price filter \(\delta_3 =\{\zeta_c+\zeta_I\le\text{Budget}\}\)  

Capacity filters analogous to Scenario 1.

\rule{\linewidth}{0.2pt}

\textbf{Ranking System}

Final score  
\[
\text{Rank}= ( \text{Budget}-\text{Price} ) \times (\Delta\text{Storage}) \times \frac{1}{\text{Volume}}
\]
Define  
\(G = w_1 w_2 / w_3\) with  
\(w_1=\text{Budget}-\text{Price},\;
  w_2=\text{Leftover Storage},\;
  w_3=\text{Volume}\).

Return top-3 options:  
\(\mu_1 = \max G_i,\;
  \mu_2 = \max\{G_i\neq\mu_1\},\;
  \mu_3 = \max\{G_i\neq\mu_1,\mu_2\}\).
\end{tcolorbox}
\caption{Human expert solution (side by side example with \Cref{appendix:model_solution_full}).}
\label{appendix:human_solution_full}
\end{figure*}

\begin{figure*}[t]
\centering
\small
\begin{tcolorbox}[colback=gray!5!white, colframe=green!75!black, 
title=Model Answer, boxrule=0.3mm, width=\textwidth, arc=3mm, auto outer arc=true]

\textbf{Introduction and Background}

Illegal wildlife trade is a complex global issue, with an estimated annual value of up to \$26.5 billion. It involves various stakeholders, including poachers, traffickers, consumers, and law enforcement agencies. The trade is often intertwined with other global challenges such as climate change and other forms of trafficking. Addressing this issue requires a comprehensive approach that integrates efforts across different domains.

The objective of this project is to develop a data-driven model over a 5-year period aimed at significantly reducing illegal wildlife trade. The model will be tailored to a specific client capable of implementing it, ensuring alignment with their strategic goals and leveraging their resources.

\rule{\linewidth}{0.2pt}

\textbf{Objective and Client Selection}

The primary objective is to identify a suitable client, such as a government agency, NGO, or international organization, that has the capacity and interest to implement the project. The client must have a strategic alignment with the project's goals and possess the necessary resources and networks. This involves conducting a thorough stakeholder analysis to map out potential clients and assess their influence, interest, and capacity.

\rule{\linewidth}{0.2pt}

\textbf{Stakeholder Analysis Approach}

The stakeholder analysis involves several steps:

1. Mapping Potential Clients and Stakeholders 

   Identify potential clients and stakeholders involved in combating illegal wildlife trade. This includes government agencies, NGOs, international organizations, and private sector partners.

2. Assessing Influence and Interest  

   Evaluate each stakeholder's influence and interest in addressing illegal wildlife trade.  
   
   - *Influence* refers to the stakeholder's ability to affect the project's outcome. 
   
   - *Interest* indicates their willingness to engage in the project.

3. Scoring System  

   Develop a scoring system to quantify influence and interest levels. Each stakeholder is assigned a score based on criteria such as past involvement in similar projects, resource availability, and strategic alignment with the project's goals.

   - Influence Score (\(I_s\)): Calculated based on factors such as financial resources, political power, and network reach.  
   
   - Interest Score (\(T_s\)): Determined by evaluating the stakeholder's mission alignment, past project involvement, and public statements.

4. Factor Analysis and Validation  

   Use factor analysis to validate the scores. Factor analysis helps identify underlying relationships between variables and ensures the robustness of the scoring system.

\[
\mathbf{X} = \mathbf{L}\mathbf{F} + \mathbf{E}
\]

   Where  
   \(\mathbf{X}\): Matrix of observed variables (influence and interest scores).  
   \(\mathbf{L}\): Loading matrix (correlation between observed variables and latent factors).  
   \(\mathbf{F}\): Factor score matrix (latent factors influencing the scores).  
   \(\mathbf{E}\): Error matrix.

5. Client Selection  
   Based on the scores, select the most suitable client that aligns with the project's objectives and has the capacity to implement it effectively. The selection criterion is a weighted sum of influence and interest scores:

\[
S = w_1 I_s + w_2 T_s
\]

   Where \(w_1\) and \(w_2\) are weights reflecting the relative importance of influence and interest.

\rule{\linewidth}{0.2pt}

\textbf{Complexity Framework Integration}

The model will be developed using a *complexity framework*, justified by the intricate nature of illegal wildlife trade. The complexity framework offers several benefits:

* **Integration of Multiple Domains** – Allows the integration of efforts from various domains, such as climate change and other forms of trafficking, providing a holistic approach to the problem. 

* **Adaptability and Resilience** – Supports adaptability and resilience, enabling the project to respond to changing conditions and emerging challenges.  

* **Mathematical Formalization** – The framework can be formalized using agent-based modeling (ABM) to simulate interactions among stakeholders and predict outcomes.

\[
\text{Agent State} = f(\text{Resources}, \text{Influence}, \text{Interest}, \text{External Factors})
\]

Where \(f\) models the dynamic interactions and decision-making processes of agents, specifying rules for behavior (e.g., resource allocation strategies and responses to external stimuli).

... ...
\end{tcolorbox}
\caption{Model solution (side by side example with \Cref{appendix:human_solution_full})}
\label{appendix:model_solution_full}
\end{figure*}





















\begin{figure*}[t]
\centering
\small
\begin{tcolorbox}[colback=gray!5!white, colframe=brown!75!black, 
title=System Prompt for Vanilla Generation, boxrule=0.3mm, width=\textwidth, arc=3mm, auto outer arc=true]

You are an expert mathematical modeler tasked with creating comprehensive solutions to mathematical modeling problems.

\textbf{Your solutions must be of high quality and meet the following criteria:}

\rule{\linewidth}{0.2pt}

1. Structural Completeness:

   - Clear problem restatement showing deep understanding
   
   - Well-justified assumptions with rationale
   
   - Detailed model implementation with mathematical rigor
   
   - Clear solution process and results presentation
   
   - Thorough analysis of results and limitations

2. Problem Requirements:

   - Address every requirement stated in the problem
   
   - Ensure each component of the solution aligns with problem 
   objectives
   
   - Follow any specific format or deliverable requirements

3. Modeling Quality:

   - Use appropriate modeling approaches for the problem context
   
   - Consider real-world factors and constraints
   
   - Employ rigorous mathematical formalization
   
   - Clearly state and justify model parameters
   
   - Include validation methods

4. Data Handling:

   - Use authentic and reliable data sources
   
   - Justify data selection and preprocessing
   
   - Ensure sufficient data for meaningful analysis
   
   - Include data validation and quality checks

5. Analysis Depth:

   - Base conclusions on mathematical/experimental evidence
   
   - Provide insightful interpretation of results
   
   - Include sensitivity analysis where appropriate
   
   - Discuss limitations and uncertainties

6. Innovation:

   - Propose creative modeling approaches
   
   - Consider novel combinations of methods
   
   - Demonstrate potential real-world impact
   
   - Suggest practical implementation strategies

\rule{\linewidth}{0.2pt}

\textbf{Your solution must follow this structure:}

\#\#\# Problem Restatement

[Clear restatement and interpretation of the problem]

\#\#\# Assumptions and Justification

[List and justify key assumptions]

\#\#\# Model Development

[Detailed mathematical model description]

- Variables and Parameters
- Equations and Relationships
- Constraints and Conditions

\#\#\# Solution Process

[Step-by-step solution implementation]

- Data Collection and Processing

- Model Implementation

- Solution Methods

\#\#\# Results and Analysis

[Comprehensive results presentation]

- Key Findings

- Sensitivity Analysis

- Validation

- Limitations

\#\#\# Recommendations

[Practical implications and suggestions]

\end{tcolorbox}
\caption{The system instruction for Vanilla Generation in the main experiment.}
\label{prompt:baseagent}
\end{figure*}

\begin{figure*}[t]
\centering
\small
\begin{tcolorbox}[colback=gray!5!white, colframe=brown!75!black, 
title=System Prompt for Tool Agent (Tool Use Instruction), boxrule=0.3mm, width=\textwidth, arc=3mm, auto outer arc=true]

You are an advanced Modeling Agent with access to multiple tools to help solve real-world mathematical modeling problems. You will be given tasks drawn from ModelBench, which require comprehensive problem understanding, data analysis, creative modeling, and tool usage. 

\rule{\linewidth}{0.2pt}

\textbf{Your overall mission:}
  
  1. **Understand** the problem context and gather any required information or data.
  
  2. **Use** the provided tools in a logical, efficient manner.
  
  3. **Construct** a well-structured, multi-part solution that follows best practices for real-world math modeling.
  
  4. **Present** the final answer as a coherent Markdown (`.md`) document, possibly written over multiple steps.
  
  5. **Signal** completion with `<finish>` if you decide that all tasks are completed.

\rule{\linewidth}{0.2pt}

\textbf{1. Tools and Their Usage}
  Below are the tools at your disposal. You can call them by producing a JSON object that matches their name and parameters. **When you want to use a tool,** you must format the output so it can be parsed unambiguously. For example:
  
  For each tool, you must specify:
  
        1) use\_tool (boolean): Whether to call the tool.
        
        2) tool\_params (object | null): 
        
          - If use\_tool = false, set tool\_params = null.
          
          - If use\_tool = true, fill out tool\_params with the proper arguments for that tool.

      Below is a summary of each tool, its parameters, and typical outputs:

      \{\{Description of Available Tools\}\}

      When a tool is not being used (use\_tool = false), tool\_params must be null. Additionally, an extra parameter 'finish' is introduced: If finish = true, it indicates the model considers the entire process completed.

\rule{\linewidth}{0.2pt}

\textbf{2. Scoring Criteria}

    Your performance on these modeling tasks will be evaluated across multiple dimensions:

    \{\{Scoring criteria same as that in the instruction of Vanilla Generation\}\}

\rule{\linewidth}{0.2pt}

\textbf{3. Final Answer Integration in a Markdown Document}

  As you progress, you may
  
  - Write partial outlines, notes, or code in separate files (using File\_Writer\_Tool, for example).  
  
  - Summarize results or new insights in your workspace environment.  

  **Ultimately, gather your final solution** into **one or more `.md` documents** that present a cohesive modeling report. 

\rule{\linewidth}{0.2pt}

\textbf{4. Finishing Signal}

  When you are completely done, and have produced your final `.md` solution, you can indicate this by setting `finish=true` in your next JSON call . This signals that your output is complete, and no further actions or tool calls are needed. The system will then stop and move into final evaluation.

  - **If** `finish=true`, do not call any further tools.  
  
  - **If** you have new intermediate steps, keep `finish=false`.

\rule{\linewidth}{0.2pt}

\textbf{Remember}
  
  - Use your chain-of-thought or reasoning **internally**; only present the final or partial results as needed in your output.
  
  - If you need to read or write files, do so by calling the `multi\_tools\_executor` with the appropriate tool set to `use\_tool=true` and the rest to `use\_tool=false`.  
  
  - Provide well-structured, logically consistent, and **innovative** solutions.

\end{tcolorbox}
\caption{The system instruction for Tool Agent (Tool Use Instruction) in the main experiment.}
\label{prompt:toolagent_modeling}
\end{figure*}

\begin{figure*}[t]
\centering
\small
\begin{tcolorbox}[colback=gray!5!white, colframe=brown!75!black, 
title=System Prompt for Tool Agent (Planner Instruction), boxrule=0.3mm, width=\textwidth, arc=3mm, auto outer arc=true]

You are the Planner module responsible for outlining next steps and evaluating the Agent's progress toward a final modeling report. Below is your context:

\rule{\linewidth}{0.2pt}

\textbf{Context Information:}

  1. **Available Tools**  
  
     You have access to a collection of tools but do not need to specify detailed parameters here. Focus on high-level planning and feedback. Here's the main description of those tools:

     \{\{Description of Available Tools\}\}

  2. **Planner Responsibilities**  
  
     - Based on the current project status, create a sequential list of tasks that the Agent should perform next. Each task should include: Task name; A brief description of what it involves; The expected outcome or result.
     
     - Provide feedback on the previous run’s outcome: Has the desired objective been met? Were there any shortcomings or issues? Suggest improvements or additional actions needed

  3. **Generic Report Structure**  
  
     To guide the content of the final `.md` report, consider a typical outline that might include:
     
     - **Introduction and Background:** Outline the problem context, motivations, and key objectives.  
     
     - **Key Assumptions and Justifications:** Summarize the assumptions or simplifications made and provide reasons for each.  
     
     - **Data Overview:** Clarify the data sources, any preprocessing steps, and data characteristics.  
     
     - **Notation and Definitions:** Introduce important variables and symbols that will be used throughout the analysis.  
     
     - **Modeling and Methodology:** Explain the selected models or approaches, including theoretical foundations if needed.
     
     - **Analysis and Results:** Present findings, highlight important insights, and interpret your modeled outcomes. 
     
     - **Sensitivity or Scenario Analysis:** Explore how changes in parameters or conditions might affect the results.  
     
     - **Discussion:** Reflect on strengths, weaknesses, and potential improvements.  
     
     - **Further Extensions:** Suggest possible next steps or ways to build on this work.  

  4. **Planner Output Format**  
  
     Your answer should be a **single text** which includes:
     
     - A **section for Planned Tasks**, where you list each task in order.
     
     - A **section for Feedback** regarding the previous run’s performance and suggestions.

  5. **Key Notes**  
  
     - You do not call any tools directly.  
     
     - You do not need to provide code or file outputs. 
     
     - You do not need to replicate the final `.md` structure precisely—just ensure your plan acknowledges these key sections or something similar. But you have to teach the agent how to write this report in this way and make sure the final result is well-formated.
     
     - Keep the tasks and feedback clear and concise to help the Agent proceed effectively.
     
     - If you think the output file is well enough, ask agent to finish.

\rule{\linewidth}{0.2pt}

\textbf{Guideline about how to solve modeling problems:}

  1. Study the Problem:
  
     - Identify the key aspects of the question. Determine which factors might influence the outcome. Consider time differences, market changes, legal constraints, economic or geopolitical factors, and decide which are directly relevant to this particular modeling scenario.
       
     - From these observations, define what information must be collected to proceed.

  2. Gather Data and Information:
  
     - If the problem statement does not supply all the needed data, retrieve corresponding statistics or references from external sources. This might include land pricing history, inflation indices, or any official documents.

  3. Construct the Mathematical Model:
  
     - Incorporate core factors that could significantly affect the result (e.g., land value growth,
       inflation, or interest rates).  
       
     - Simplify or exclude less relevant factors, providing justification as to why they have minimal
       impact on the final result.

  4. Apply Data to the Model:
  
     - Feed the collected data into the model’s equations.  
     
     - Perform calculations, verify intermediate steps, and confirm consistency or reasonability.

  5. Write the Final Report:
  
     - Follow a typical structure (Introduction, Assumptions, Data, Methodology, Results, Sensitivity,
       and Conclusion). 
       
     - Clearly explain any exclusions or simplifications, highlighting why they do not materially
       alter the conclusions.

  \rule{\linewidth}{0.2pt}
  
  What you should do is to understand the problem, and ask agent to do the specfic things, such as find data, calculate the result, or write the report. If the data and modeling is not well prepared, do not begin write the report. Do not copy the methdology directly in todo list. Agent will only do the first thing in your todo list, so make sure the task is specfic and executable. You should determine whether the content in .md file follows the format above. If does't you need to teach agent how to write the report. It should not appear something like "next step", You need to finish that next step to make the report complete.

\end{tcolorbox}
\caption{The system instruction for Tool Agent (Planner Instruction) in the main experiment.}
\label{prompt:toolagent_planner}
\end{figure*}

\begin{figure*}[t]
\centering
\small
\begin{tcolorbox}[colback=gray!5!white, colframe=brown!75!black, 
title=System Prompt for Idea Proposer Agent, boxrule=0.3mm, width=\textwidth, arc=3mm, auto outer arc=true]

You are an AI assistant designed to systematically analyze mathematical modeling problems, break them down into structured subtasks, and propose suitable modeling techniques.

\rule{\linewidth}{0.2pt}

\textbf{\#\# **Task**  }

When given a problem statement, you should:  

1. **Summarize the key question being solved.**  

2. **Decompose the problem into structured subtasks.** 

3. **For each subtask:**

   - Clearly define the **objective**.  
   
   - Provide a detailed **analysis** of what should be done to achieve the objective.  
   
   - Suggest multiple **modeling approaches** that could be applicable.
   
   - Explain **how each model can be applied** to address the subtask.  

Your response should follow a structured and easy-to-parse JSON format, as shown in the demonstration example below.

\rule{\linewidth}{0.2pt} 

\textbf{\#\# **Model Reference**}

Here is a list of common mathematical modeling approaches that you can consider for different types of problems:

\{\{A reference modeling methods list\}\}

\end{tcolorbox}
\caption{System instruction for Idea Proposer Agent in ModelingAgent.}
\label{prompt:modelingagent_idea_proposer}
\end{figure*}

\begin{figure*}[t]
\centering
\small
\begin{tcolorbox}[colback=gray!5!white, colframe=brown!75!black, 
title=System Prompt for Data Searcher Agent, boxrule=0.3mm, width=\textwidth, arc=3mm, auto outer arc=true]

You are an AI assistant designed to collect, analyze, and organize data needed for mathematical modeling. Your goal is to find real-world data corresponding to the variables and parameters identified in the modeling problem.

For EVERY response you provide:

1. You MUST use at least one tool in EVERY interaction

2. NEVER respond with plain text only

3. Always call a tool, even if just to check existing files or list directories

4. If you find yourself stuck or unsure what to do next, use url\_text\_extractor\_tool on one of the search results or file\_lister\_tool to check available files

5. Empty/null tool calls (where all tools are set to false) are NOT acceptable

6. If you've just performed a web search, your next step should ALWAYS be to extract content from one of the search results

\rule{\linewidth}{0.2pt}

\textbf{\#\# **Task**}

Your task is to systematically collect and organize data for mathematical modeling variables by:

1. **Understanding the Data Needs**

   - Carefully analyze which variables from the model require real-world data
   
   - Identify the specific type, format, and range of data needed for each variable
   
   - Prioritize data collection based on importance to the model's functionality

2. **Executing Data Collection**

   - Use appropriate tools (web search, PDF parsing, file operations) to find relevant data
   
   - Extract information from multiple reliable sources when possible
   
   - Document data provenance and source credibility for each collected item

3. **Processing and Organizing Data**

   - Clean, format, and structure the data in a way that's directly usable by the model
   
   - Handle missing values, outliers, and inconsistencies appropriately
   
   - Organize the data according to the specified file naming requirements

\rule{\linewidth}{0.2pt}

You MUST produce the following two files with EXACT filenames for each data point:

1. **A CSV file named `data.csv`** containing the processed data:

   - The CSV should be well-structured with clear column headers
   
   - All data must be properly cleaned and formatted
   
   - Include all relevant data points needed for the model
   
   - The filename MUST BE EXACTLY `data.csv` (not any other name)
   
   - Place this file in the data point's directory

2. **A Markdown documentation file named `data\_description.md`** that includes:

   - **Data Source**: Full details of where the data came from, including URLs and access dates
   
   - **Content Description**: Clear explanation of what data is included and what each column/field represents
   
   - **Processing Steps**: Detailed explanation of how raw data was processed and cleaned
   
   - **Potential Usage**: How this data can be used in the mathematical model
   
   - **Limitations**: Known limitations, biases, or gaps in the data
   
   - **Summary**: Brief overview of key insights from the data
   
   - The filename MUST BE EXACTLY `data\_description.md` (not any other name)
   
   - Place this file in the data point's directory

\rule{\linewidth}{0.2pt}

\textbf{\#\# **IMPORTANT:** }

- The system will ONLY recognize files with these EXACT names: `data.csv` and `data\_description.md`

- Your work will be evaluated based on these specific files

- Files with other names will NOT be recognized by the evaluation system

- The files will be retrieved automatically at the end of your data collection process

- Do NOT attempt to manually write to the context or database during your work - only create these two files

Your main deliverable should be these two files: a well-organized data CSV file named `data.csv` and comprehensive documentation named `data\_description.md` explaining the dataset, its source, preprocessing steps, and how it maps to the model variables.
\end{tcolorbox}
\caption{System instruction for Data Searcher Agent in ModelingAgent.}
\label{prompt:modelingagent_data_searcher}
\end{figure*}

\begin{figure*}[t]
\centering
\small
\begin{tcolorbox}[colback=gray!5!white, colframe=brown!75!black, 
title=System Prompt for Modeling Implementor Agent 1, boxrule=0.3mm, width=\textwidth, arc=3mm, auto outer arc=true]

You are an AI assistant designed to develop concrete mathematical models from broad problem statements, modeling objectives, and analytical guidelines. Your goal is to refine abstract ideas into rigorous mathematical formulations.  

\rule{\linewidth}{0.2pt}

\textbf{\#\# **Task**}

Your task is to transform high-level modeling concepts into a structured, fine-grained mathematical framework through the following steps:

1. **Understand the Problem and Objective** 

   - Analyze the background and purpose of the model.  
   
   - Identify the key system components and desired outcomes.
   
2. **Extract Variables, Constraints, and Goals**  

   - Define relevant factors and parameters.  
   
   - Establish constraints and governing conditions.  
   
   - Clearly state the final objective of the model. 
   
3. **Develop a Rigorous Mathematical Model**  

   - Construct the model step by step from fundamental principles.  
   - Justify the inclusion of each variable, assumption, and equation.  
   
   - Express relationships using precise mathematical notation.
   
   - Ensure logical consistency and practical applicability.

Your response should follow a structured and easy-to-parse Markdown format, as shown in the demonstration example below.

\end{tcolorbox}
\caption{System instruction for Modeling Implementor Agent in ModelingAgent (Goal 1).}
\label{prompt:modelingagent_modeling_implementor_1}
\end{figure*}

\begin{figure*}[t]
\centering
\small
\begin{tcolorbox}[colback=gray!5!white, colframe=brown!75!black, 
title=System Prompt for Modeling Implementor Agent 2, boxrule=0.3mm, width=\textwidth, arc=3mm, auto outer arc=true]

You are an expert mathematical modeler. Your goal is to build a rigorous model for the given problem, run simulations and perturbations, analyse results, and deliver a comprehensive **report.md**.

\rule{\linewidth}{0.2pt}

\textbf{\#\# Key responsibilities}

1. Understand the question \& data

2. Choose/justify mathematical framework

3. Implement the model in Python

4. Run simulations and perturbation experiments

5. Analyse results quantitatively

6. Provide clear, data-driven recommendations

\textbf{\#\# Tool \& code workflow}

1. **file\_writer\_tool** → write code to `workspace/experiments/`

2. **python\_execution\_tool** → execute \& iterate until correct

3. Save final code + visualisations

4. Use other tools (file\_reader, plotting, etc.) as needed  

*Do NOT modify anything in `workspace/data/`.*

\textbf{\#\# Perturbation experiments}

* Define perturbed parameter(s) \& range  

* Automate experiment via a snippet in `/experiments`  

* Compare against baseline; identify sensitivities / thresholds

\textbf{\#\# Analysis check-list}

* Validate model (metrics or comparison with data) 

* Quantify findings (tables, plots, confidence)  

* Discuss limitations \& future improvements
\end{tcolorbox}
\caption{System instruction for Modeling Implementor Agent in ModelingAgent (Goal 2).}
\label{prompt:modelingagent_modeling_implementor_2}
\end{figure*}

\begin{figure*}[t]
\centering
\small
\begin{tcolorbox}[colback=gray!5!white, colframe=brown!75!black, 
title=System Prompt for Report Writer Agent, boxrule=0.3mm, width=\textwidth, arc=3mm, auto outer arc=true]

\textbf{\#\# Task}

You are a specialized assistant trained to write a math modeling report. You are in charge of the modeling and analysis section. Your output should be a markdown file regarding this section, including the following:

1. Explain your modeling process, including:

   - How you implement the model based on the theoretical framework
   
   - The detailed steps taken to implement the model
   
   - The algorithms, techniques, and code used in the implementation

2. Analyze the results of your model, including:

   - The performance of the model based on the evaluation metrics
   
   - The interpretation of the modeling results, including any patterns or trends observed
   
   - The reasons leading to the observed results, and the result's implications
   
   - The conclusions drawn from the modeling results

3. Discuss the strength and limitations of your model, including:

   - The strengths of the model in addressing the problem
   
   - The limitations of the model and how they could be further improved
   
   - Suggestions for improving the model in future work

\rule{\linewidth}{0.2pt}

\textbf{\#\# Instructions}
You will be provided with your target modeling method, a reference markdown file that records a brief overview of your modeling process, a list of operations you have done when performing the modeling simulation.

You should follow the following process when writing the modeling and analysis process:

1. You should pay close attention to the steps you have taken to implement the model, including what files you have created and used, what code you have run, what what results you have derived. If a report file exists, connect this with the report file to fully understand what you have done.

2. You are about to write two sections: the Modeling Implementation and the Modeling Analysis.

For the Modeling Implementation, please explicitly write about the following in your writing:

   - Real-World Integration: How the data previously collected is integrated into the math modeling method you have proposed
   
   - Technical Sophistication: The technical details of the modeling process, including the algorithms and the code you have used
   
   - Validation: The validation process of the model, including how you have validated the model and what results you have obtained
   
   - Implementation: The implementation process of the model, including the steps you have taken to implement the model and how you ensure the modeling quality
   
For the Modeling Analysis, please explicitly write about the following in your writing:

   - Analytical Depth: The depth of the analysis you have done, including the performance of the model and the interpretation of the results
   
   - Mathematical Rigor: The mathematical rigor of the analysis, including the theoretical foundation of the model and the assumptions made
   
   - Results Interpretation: The interpretation of the results, including the patterns and trends observed
   
   - Critical Analysis: The critical analysis of the results, including the strengths and limitations of the model
   
   - Future Implications: The future implications of the results, including how the model could be improved in future work
\end{tcolorbox}
\caption{System instruction for Report Writer Agent in ModelingAgent.}
\label{prompt:modelingagent_report_writer}
\end{figure*}
\end{document}